\documentclass[11pt]{article}

\usepackage[T1]{fontenc}
\usepackage[utf8]{inputenc}
\usepackage{lmodern}
\usepackage{textcomp}

\DeclareUnicodeCharacter{00B0}{\textdegree}
\DeclareUnicodeCharacter{00B2}{\textsuperscript{2}}
\DeclareUnicodeCharacter{00D7}{\ensuremath{\times}}
\DeclareUnicodeCharacter{0394}{\ensuremath{\Delta}}
\DeclareUnicodeCharacter{2013}{--}
\DeclareUnicodeCharacter{2019}{'}
\DeclareUnicodeCharacter{201C}{``}
\DeclareUnicodeCharacter{201D}{''}
\DeclareUnicodeCharacter{2192}{\ensuremath{\rightarrow}}
\DeclareUnicodeCharacter{21D2}{\ensuremath{\Rightarrow}}
\DeclareUnicodeCharacter{2248}{\ensuremath{\approx}}

\usepackage[margin=1in]{geometry}
\usepackage{amsmath,amssymb}
\usepackage{graphicx}
\setkeys{Gin}{width=\linewidth,height=0.9\textheight,keepaspectratio}
\usepackage{booktabs}
\usepackage{longtable}
\usepackage{array}
\usepackage{float}
\usepackage{caption}
\usepackage{xcolor}
\usepackage{hyperref}
\usepackage{calc}

\hypersetup{
  colorlinks=true,
  linkcolor=blue,
  citecolor=blue,
  urlcolor=blue
}

\providecommand{\tightlist}{%
  \setlength{\itemsep}{0pt}\setlength{\parskip}{0pt}}

\newlength{\cslhangindent}
\newlength{\csllabelwidth}
\newenvironment{CSLReferences}[2] % #1 hanging indent, #2 entry spacing
  {\setlength{\parindent}{0pt}%
   \setlength{\parskip}{#2\baselineskip}%
   \ifodd #1\relax
     \setlength{\leftskip}{\cslhangindent}%
     \setlength{\parindent}{-\cslhangindent}%
   \fi}
  {}

\usepackage{color}
\usepackage{fancyvrb}

\DefineVerbatimEnvironment{Highlighting}{Verbatim}{commandchars=\\\{\}}
\newenvironment{Shaded}{}{}

\newcommand{\NormalTok}[1]{#1}

\title{Modality-Driven Search with Holistic Trace Judging for ARC-AGI-2}
\author{Johan
Land \\ Independent Researcher \\ \texttt{johan.land@gmail.com}}
\date{2026-02-03}

\begin{document}
\maketitle

\begin{abstract}
Large language models can produce fluent, internally coherent reasoning
traces for abstract reasoning tasks while still being confidently wrong
--- making selection among candidates, not just generation, the central
challenge. I present a solver for ARC-AGI-2 (a few-shot visual reasoning
benchmark) built around two principles: (i) treating reasoning
modalities as search operators, generating diverse candidates
independently across text, image, and code channels, and (ii)
context-preserving holistic judging, in which a judge model jointly
compares all candidate reasoning traces within a single long-context
prompt. Unlike self-consistency or majority voting, this approach
reliably recovers correct minority hypotheses on tasks where the modal
answer is wrong.

On the ARC Prize semi-private evaluation set, the solver achieves 72.9\%
at \$38.99/task --- the highest score on the verified leaderboard at the
time of writing, exceeding the best standalone frontier models (GPT-5.2
Pro at 54.2\%, Gemini 3 Pro at 54.0\%) by +18.7 percentage points. On
the public evaluation set, it achieves 76.1\% at \$19.69/task. I release
the full source code and document extensive negative results, including
the finding that prescriptive prompting templates and iterative
refinement systematically reduce hypothesis diversity and degrade
performance.
\end{abstract}

\hypertarget{introduction}{%
\section{Introduction}\label{introduction}}

A central challenge in applying LLMs to abstract reasoning is not just
producing candidate solutions, but \textbf{knowing what is right and
what is wrong} in a setting where models can be confidently
incorrect---even when they provide detailed, plausible reasoning traces.

ARC-AGI-2 was designed to be \emph{easy for humans and hard for AI},
and---critically---to measure both \textbf{capability} and
\textbf{efficiency} (cost). Progress on ARC-style benchmarks has also
been rapid rather than stagnant: ARC Prize reports significant
year-over-year improvements driven by frontier ``reasoning systems'' and
application-layer refinement harnesses (Chollet et al. 2024).

This paper describes an approach that treats \textbf{modalities as
search operators} and uses \textbf{judging as the final selection
mechanism}: generate diverse candidate solutions across independent
reasoning channels, then select among them using context-preserving
holistic judging.\footnote{This work has not been peer-reviewed and is
  intended as a technical preprint.}

\hypertarget{contributions}{%
\subsection{Contributions}\label{contributions}}

\begin{itemize}
\tightlist
\item
  \textbf{A modality-driven search solver} that generates candidates
  independently across text, image, and code reasoning channels to
  produce diverse candidate solutions.
\item
  \textbf{A context-preserving holistic judge} that reads all candidate
  traces jointly to select the best outputs. Unlike standard
  self-consistency (majority vote) or per-candidate scoring, this judge
  identifies correct \emph{minority} hypotheses by comparing full
  reasoning traces in a single context window --- yielding +7 solved
  instances over majority vote at only 13\% of total system cost
  (Section 7).
\item
  \textbf{Verified ARC-AGI-2 semi-private performance:} 72.9\% at
  \$38.99/task on the ARC Prize Verified leaderboard dataset\footnote{https://arcprize.org/leaderboard}
  --- the highest score on the leaderboard at the time of writing,
  exceeding the best standalone frontier models (GPT-5.2 Pro at 54.2\%,
  Gemini 3 Pro at 54.0\%) by +18.7 percentage points.
\item
  \textbf{Public eval performance:} 76.11\% at \$19.69/task
  (self-measured).
\item
  \textbf{AI-assisted development:} the solver was developed with AI
  assistance for both implementation and architectural design (Section
  3.3).
\item
  \textbf{Open-source release} of the full source code\footnote{https://github.com/beetree/ARC-AGI}
  plus detailed negative results documenting what did not work and why.
  The complete public-evaluation run data --- over 7 million lines of
  prompts, responses, reasoning traces, and judge transcripts --- is
  also released\footnote{https://www.kaggle.com/code/johanland/johan-land-solver-v7-public/comments?scriptVersionId=290052212}
  to support future research.
\end{itemize}

\begin{center}\rule{0.5\linewidth}{0.5pt}\end{center}

\hypertarget{background-and-related-work}{%
\section{Background and Related
Work}\label{background-and-related-work}}

This paper sits at the intersection of (i) \textbf{abstraction-centric
few-shot generalization} as instantiated by ARC-style tasks, (ii)
\textbf{search-based and neuro-symbolic solvers} that treat ARC as a
latent-program induction problem, and (iii) \textbf{test-time compute
scaling} strategies for frontier LLMs---especially approaches that
generate multiple candidate trajectories and then \textbf{select,
verify, or judge} among them. This section reviews the ARC/ARC-AGI
benchmark lineage, highlights major families of solver approaches, and
situates ``modality search + trace-preserving holistic judging''
relative to the most relevant prior work.

\hypertarget{arc-and-arc-agi-as-a-benchmark-for-abstraction-under-minimal-priors}{%
\subsection{ARC and ARC-AGI as a benchmark for abstraction under minimal
priors}\label{arc-and-arc-agi-as-a-benchmark-for-abstraction-under-minimal-priors}}

The Abstraction and Reasoning Corpus (ARC) was introduced by Chollet
(2019) as part of a broader argument for measuring intelligence as
\textbf{skill-acquisition efficiency}---how effectively a system can
acquire and apply new skills under constrained experience and priors.
ARC's design emphasizes rapid generalization from a small number of
examples, with tasks intended to require only relatively elementary
``core knowledge'' and to discourage reliance on domain knowledge or
internet-scale memorization.

ARC tasks are framed as \textbf{few-shot input--output induction}: given
a handful of training demonstrations (pairs of grids), the solver must
infer an underlying transformation rule and apply it to a held-out test
input. The hallmark difficulty is \textbf{underspecification}: multiple
hypotheses can explain the training pairs, but only a subset will
transfer to the test instance, so solvers must cope with an intense
``many consistent hypotheses'' regime where superficial fit is not
enough.

As a concrete example, Figure 1 shows all three training pairs and the
test input from task \texttt{3dc255db}.\footnote{https://arcprize.org/play?task=3dc255db}
A human might interpret the shapes as ``spaceships'': colored particles
sit inside each ship on the exhaust side, and the transformation removes
them from the interior and places them on the nose, extending the ship
in its direction of travel. The solver must infer this rule ---
identifying containment, directionality, and the interior/exterior
distinction --- from only three training demonstrations, then apply it
to the unseen test input (bottom row). This task remains unsolved by the
solver described in this paper: all 29 candidates failed, and none of
GPT-5.2, Gemini 3, or Opus 4.5 produced a correct output.

\begin{figure}
\centering
\includegraphics{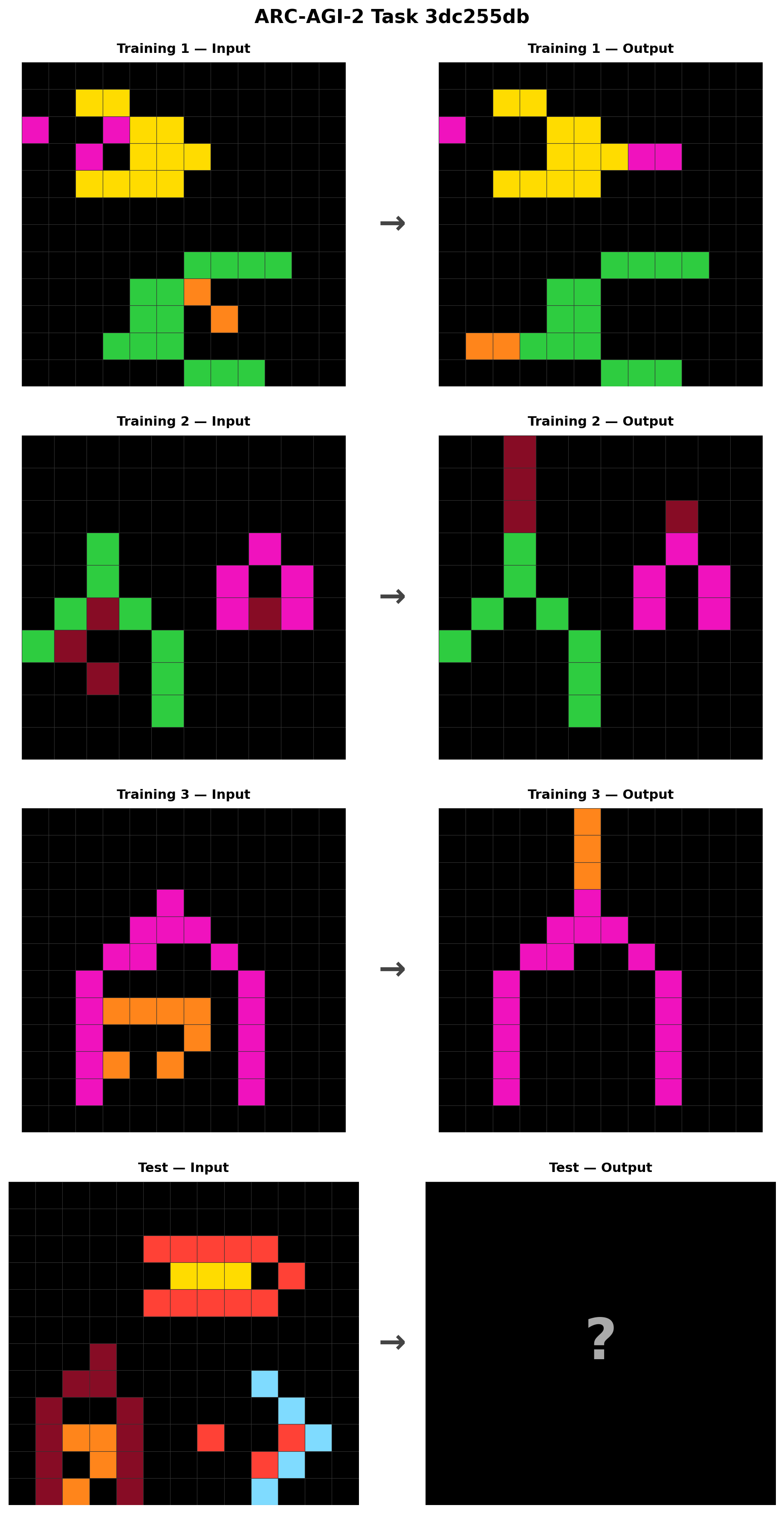}
\caption{ARC-AGI-2 task \texttt{3dc255db}. A human might see
``spaceships'' with particles on the exhaust side. The transformation
removes the particles from the interior and extends them from the nose.
Three training pairs (rows 1--3) demonstrate the rule; the test input
(row 4) must be solved from these examples alone. This task remains
unsolved.}
\end{figure}

A recurring theme in ARC research is that the benchmark stresses
\textbf{compositional abstraction} (e.g., combining multiple latent
concepts) and \textbf{distribution shift within each task} (training
vs.~test), rather than data-driven interpolation across a large IID
dataset. This is part of what makes ARC resistant to straightforward
deep learning approaches trained on the released tasks alone, and it
motivates solver families that include explicit search, symbolic
representations, or test-time adaptation.

\hypertarget{the-arc-prize-ecosystem-and-the-evolution-to-arc-agi-2}{%
\subsection{The ARC Prize ecosystem and the evolution to
ARC-AGI-2}\label{the-arc-prize-ecosystem-and-the-evolution-to-arc-agi-2}}

\hypertarget{arc-competitions-and-the-slow-progress-era}{%
\subsubsection{ARC competitions and the ``slow progress''
era}\label{arc-competitions-and-the-slow-progress-era}}

After ARC's release, the first major public competition was the Kaggle
``Abstraction and Reasoning Challenge'' (2020). The best-performing
solutions in that era were largely \textbf{program-synthesis /
DSL-search systems}, and performance improved only gradually for several
years---an arc that ARC Prize reports explicitly document when
motivating why new benchmark design was needed.

The ARC Prize effort expanded this ecosystem with additional competitive
events and a more formalized reporting and verification posture,
including an explicit policy around ``ARC Prize Verified'' scores to
reduce confusion arising from incomparable, self-reported results.

\hypertarget{arc-agi-2-design-goals-splits-and-evaluation-protocol}{%
\subsubsection{ARC-AGI-2: design goals, splits, and evaluation
protocol}\label{arc-agi-2-design-goals-splits-and-evaluation-protocol}}

ARC-AGI-2 was introduced as a second-generation benchmark intended to
provide a more informative signal at the frontier of reasoning systems.
The technical report (Chollet et al. 2024) highlights several goals:
maintain the original ARC principles and format, reduce susceptibility
to brute-force program search, incorporate \textbf{first-party human
testing}, and increase ``signal bandwidth'' (a wider useful range of
scores to track progress).

ARC-AGI-2 also formalizes dataset splits and calibration:

\begin{itemize}
\tightlist
\item
  \textbf{Training set}: 1000 public tasks spanning a wide range of
  difficulties.
\item
  \textbf{Public evaluation set}: 120 calibrated public tasks.
\item
  \textbf{Semi-private evaluation set}: 120 calibrated non-public tasks
  (used for the live leaderboard and ARC Prize leaderboard).
\item
  \textbf{Private evaluation set}: 120 calibrated non-public tasks (used
  for final contest ranking).
\end{itemize}

A key protocol detail is the use of \textbf{pass@2} scoring,
acknowledging that some tasks can contain genuine ambiguity; ARC Prize
materials emphasize that the benchmark's human calibration also used the
same ``two attempts'' framing (e.g., tasks solved pass@2 by at least two
humans).

Finally, ARC Prize's public-facing evaluation culture emphasizes not
just raw accuracy but also \textbf{efficiency} and comparability
(including leaderboard reporting and verification norms).

\hypertarget{classical-arc-solvers-dsl-program-synthesis-and-enumerative-search}{%
\subsection{Classical ARC solvers: DSL program synthesis and enumerative
search}\label{classical-arc-solvers-dsl-program-synthesis-and-enumerative-search}}

The most historically influential ``classical'' ARC solver family treats
tasks as \textbf{latent programs} composed from a hand-designed library
of primitives (a DSL). The Kaggle 2020 top solutions are widely
recognized as belonging to this category: they relied on enumerating
candidate transformation chains (sometimes aggressively optimized in
low-level languages) to find programs consistent with the
demonstrations, and then applying the discovered program to the test
input.

These approaches matter as baseline ``proofs of tractability'' for a
subset of ARC tasks and as a reminder that \textbf{search can compensate
for weak learned priors}---but they also illuminate why ARC-AGI-2
explicitly tries to be ``less brute-forcible.'' In particular, ARC
Prize's ARC-AGI-2 announcement and technical report describe removing or
redesigning evaluation tasks that were overly susceptible to brute-force
search methods.

Beyond DSL enumeration, related symbolic traditions include approaches
that use \textbf{compression / minimum description length (MDL)}
principles to guide search over explanations. Ferré (2021) provides a
representative example: describing grids using explicit models and
searching for explanations that compress the observations, illustrating
an alternative axis of ``prior + search'' design that emphasizes
interpretability and parsimony rather than only brute-force enumeration.

\hypertarget{benchmark-extensions-and-data-augmentation-in-the-arc-domain}{%
\subsection{Benchmark extensions and data augmentation in the ARC
domain}\label{benchmark-extensions-and-data-augmentation-in-the-arc-domain}}

Several efforts address limitations of the original ARC release:
ConceptARC (Moskvichev, Odouard, and Mitchell 2023) introduces
concept-grouped task variants to probe systematic generalization;
ARC-GEN (Moffitt 2025) proposes a procedural benchmark generator; and
Re-ARC (Hodel 2024) provides a programmatic reproduction of ARC tasks
frequently used for generating synthetic training variations.

\hypertarget{learned-approaches-transduction-induction-and-hybridization}{%
\subsection{Learned approaches: transduction, induction, and
hybridization}\label{learned-approaches-transduction-induction-and-hybridization}}

ARC has also been attacked from a purely learned perspective (e.g.,
treating ARC as image-to-image translation), but results historically
lagged behind symbolic-search systems and humans. ARC Prize's technical
reporting explicitly notes early deep-learning baselines performing very
poorly on ARC-AGI, motivating hybrid approaches and new data strategies.

Recent work has sharpened the conceptual distinction between:

\begin{itemize}
\tightlist
\item
  \textbf{Induction}: infer a latent function/program consistent with
  demonstrations, then apply it.
\item
  \textbf{Transduction}: directly predict the test output conditioned on
  the demonstrations and test input, without explicitly representing a
  latent program.
\end{itemize}

Li et al. (2025) study this tradeoff on ARC and find induction and
transduction succeed on different problem families; importantly, they
show that combining (ensembling) these paradigms can approach
human-level performance on the original ARC benchmark under their
experimental setup and synthetic training regime. Complementary work on
small or specialized models (Fletcher-Hill 2024; Puget 2024) explores
architectural inductive biases for 2D grid structure --- highlighting
the increasing importance of \textbf{test-time procedures} (refinement,
adaptation, search) even when the base model is learned.

\hypertarget{test-time-adaptation-and-compute-scaling-for-arc-style-tasks}{%
\subsection{Test-time adaptation and compute scaling for ARC-style
tasks}\label{test-time-adaptation-and-compute-scaling-for-arc-style-tasks}}

A major recent shift in ARC solving is the move from purely ``static''
models or solvers to methods that treat each ARC task as an opportunity
for \textbf{test-time learning} or \textbf{test-time search}.

Akyürek et al. (2025) demonstrate that updating model parameters at test
time (under carefully controlled procedures) can yield surprisingly
strong gains on ARC-like reasoning, reinforcing the idea that ARC is a
testbed for \emph{within-task} adaptation rather than only pretraining.

Other ARC Prize--era work explores search and induction at test time in
more explicitly programmatic spaces: Macfarlane and Bonnet (2025)
combines learned representations with explicit search over programs,
while Ouellette (2024) compares neurally-guided program induction
paradigms across grid, program, and transform spaces. At the systems
level, competitive solvers increasingly resemble \textbf{pipelines} that
integrate synthetic data, model adaptation, search components, and
ensembles (Chollet et al. 2024).

\hypertarget{llm-era-reasoning-generating-and-coordinating-multiple-trajectories}{%
\subsection{LLM-era reasoning: generating and coordinating multiple
trajectories}\label{llm-era-reasoning-generating-and-coordinating-multiple-trajectories}}

Independently of ARC, the broader LLM literature has developed a family
of \textbf{test-time compute} and \textbf{trajectory diversification}
techniques that are directly relevant to the solver architecture in this
paper. Snell et al. (2025) show that optimally allocating test-time
compute --- e.g., by generating and selecting among multiple candidate
solutions --- can be more effective than scaling model parameters,
providing formal grounding for architectures that trade inference-time
search budget for performance.

\hypertarget{chain-of-thought-and-sampling-based-diversification}{%
\subsubsection{Chain-of-thought and sampling-based
diversification}\label{chain-of-thought-and-sampling-based-diversification}}

Chain-of-thought prompting (Wei et al. 2022) established that eliciting
intermediate reasoning steps can improve performance on multi-step
tasks.

Self-consistency (Wang et al. 2023) then proposed a simple but
influential extension: sample multiple reasoning paths and select the
most consistent final answer, demonstrating that \emph{diversity +
aggregation} can outperform a single greedy reasoning trace.

\hypertarget{reasoning-as-search-over-thoughts}{%
\subsubsection{Reasoning as search over
``thoughts''}\label{reasoning-as-search-over-thoughts}}

Tree of Thoughts {[}ToT; Yao, Yu, et al. (2023){]} reframes inference as
explicit search over intermediate ``thoughts,'' enabling branching,
lookahead, and backtracking. Graph of Thoughts {[}GoT; Besta et al.
(2024){]} generalizes this idea to arbitrary graph-structured reasoning
artifacts, emphasizing more flexible dependency structures across
intermediate units of information.

These frameworks provide language for understanding ARC solvers that
``branch'' over hypotheses rather than commit early, and they motivate
treating candidate generation as a search process rather than a single
pass.

\hypertarget{tool-use-and-program-aided-reasoning}{%
\subsubsection{Tool use and program-aided
reasoning}\label{tool-use-and-program-aided-reasoning}}

ReAct (Yao, Zhao, et al. 2023) interleaves reasoning traces with actions
(tool calls / environment interactions), highlighting how external tools
can mitigate hallucination and enable more reliable task completion.

Toolformer (Schick et al. 2023) provides a training-time perspective,
showing that LMs can learn to decide \emph{when and how} to call tools
via self-supervision, further legitimizing tool-augmented reasoning as a
general capability axis.

PAL {[}Program-aided Language Models; Gao et al. (2023){]} formalizes a
closely related idea: use the LM to translate problems into runnable
code, then outsource exact computation to an interpreter---often
improving accuracy on tasks where arithmetic/logic errors dominate.

These ideas are directly relevant to ARC, where code synthesis can serve
both as a hypothesis generator and as a way to expose structured
intermediate artifacts (programs, tool outputs) that may be useful
during downstream selection.

\hypertarget{iterative-refinement-and-memory-based-improvement}{%
\subsubsection{Iterative refinement and memory-based
improvement}\label{iterative-refinement-and-memory-based-improvement}}

Self-Refine (Madaan et al. 2023) shows that a single LM can iteratively
improve its output by generating feedback and revisions in a loop,
without gradient updates. Reflexion (Shinn et al. 2023) similarly aims
to improve test-time performance by incorporating feedback into a memory
buffer that influences subsequent attempts, framing improvement as
``verbal reinforcement learning'' rather than weight updates.

These methods relate to ARC attempts that ``try multiple times'' and
learn from earlier mistakes, though they also introduce a key trade-off:
iterative refinement can increase compute while risking
\textbf{anchoring} to early hypotheses---an issue that becomes acute on
ARC tasks where early commitments can be misleading.

\hypertarget{selection-verification-and-llm-as-a-judge-paradigms}{%
\subsection{Selection, verification, and ``LLM-as-a-judge''
paradigms}\label{selection-verification-and-llm-as-a-judge-paradigms}}

Generating diverse candidates is only half the problem; the other half
is selecting among them. In the LLM ecosystem, selection is increasingly
delegated to learned or LLM-based evaluators, giving rise to the
``LLM-as-a-judge'' paradigm.

Zheng et al. (2023) formalized and stress-tested LLM judging in the
context of MT-Bench and Chatbot Arena, showing that strong LLM judges
can correlate well with human preferences while also exhibiting
systematic biases (e.g., position and verbosity biases). Subsequent work
has explored structured multi-agent evaluation (judge-and-jury designs)
and cautioned that judging format (pairwise vs.~pointwise, aggregation
schemes) materially affects robustness.

For ARC in particular, selection is unusually difficult because many
hypotheses fit the demonstrations yet fail on the test instance. A good
judge must reward \textbf{transfer-valid abstractions}, not merely
fluent rationales or training-pair fit.

\hypertarget{positioning-of-this-work-within-the-landscape}{%
\subsection{Positioning of this work within the
landscape}\label{positioning-of-this-work-within-the-landscape}}

Relative to the above literature, the approach in this paper can be
viewed as a systems-level composition of three trends:

\begin{enumerate}
\def\labelenumi{\arabic{enumi}.}
\item
  \textbf{Hypothesis generation as explicit search} (a lineage shared
  with DSL/program-synthesis solvers and ToT/GoT-style inference (Yao,
  Yu, et al. 2023; Besta et al. 2024)), but broadened beyond a single
  representation to \emph{multiple reasoning modalities}.
\item
  \textbf{Tool- and program-mediated reasoning} (Yao, Zhao, et al. 2023;
  Gao et al. 2023; Schick et al. 2023) used not just for execution but
  also as a way to produce richer intermediate artifacts that can be
  consumed by downstream selection.
\item
  \textbf{Judge-based selection} {[}LLM-as-a-judge; Zheng et al.
  (2023){]} adapted from general LLM evaluation into an in-task
  meta-reasoning component, with the additional complication that ARC
  demands judging \emph{generalization under underspecification}, not
  merely surface quality.
\end{enumerate}

What is relatively distinctive in the ARC context is the combination of
(i) \textbf{heterogeneous candidate generators} (text, code, visual,
extended deliberation) and (ii) \textbf{context-preserving comparison
over full traces}, rather than scalar scoring or consensus
compression---an axis that is motivated both by historical ARC solver
failure modes (overfitting via brittle heuristics) and by known
limitations of LLM judges under compressed, bias-prone evaluation
formats.

\begin{center}\rule{0.5\linewidth}{0.5pt}\end{center}

\hypertarget{method-modality-driven-search-and-architecture}{%
\section{Method: Modality-Driven Search and
Architecture}\label{method-modality-driven-search-and-architecture}}

\hypertarget{core-idea-independent-candidates-across-modalities-maximize-diversity}{%
\subsection{Core idea: independent candidates across modalities maximize
diversity}\label{core-idea-independent-candidates-across-modalities-maximize-diversity}}

The solver is built around a practical observation: to solve tasks that
frontier systems and labs \emph{do not already solve}, the correct
solution is often a \textbf{minority hypothesis}. If ``the most common
solution'' were correct, the task would likely already be within the
main cluster of model behavior.

Therefore, the solver's first phase intentionally creates many
\emph{independent} candidate solutions across heterogeneous reasoning
modalities (text, image, and code), maximizing the probability that at
least one candidate captures a genuinely novel hypothesis. Each modality
provides a structurally different representation of the same task, which
empirically produces candidates that cluster differently (Section 6.5).

\hypertarget{pipeline-overview}{%
\subsection{Pipeline overview}\label{pipeline-overview}}

Figure 2 shows the end-to-end pipeline. For each ARC task:

\begin{figure}
\centering
\includegraphics{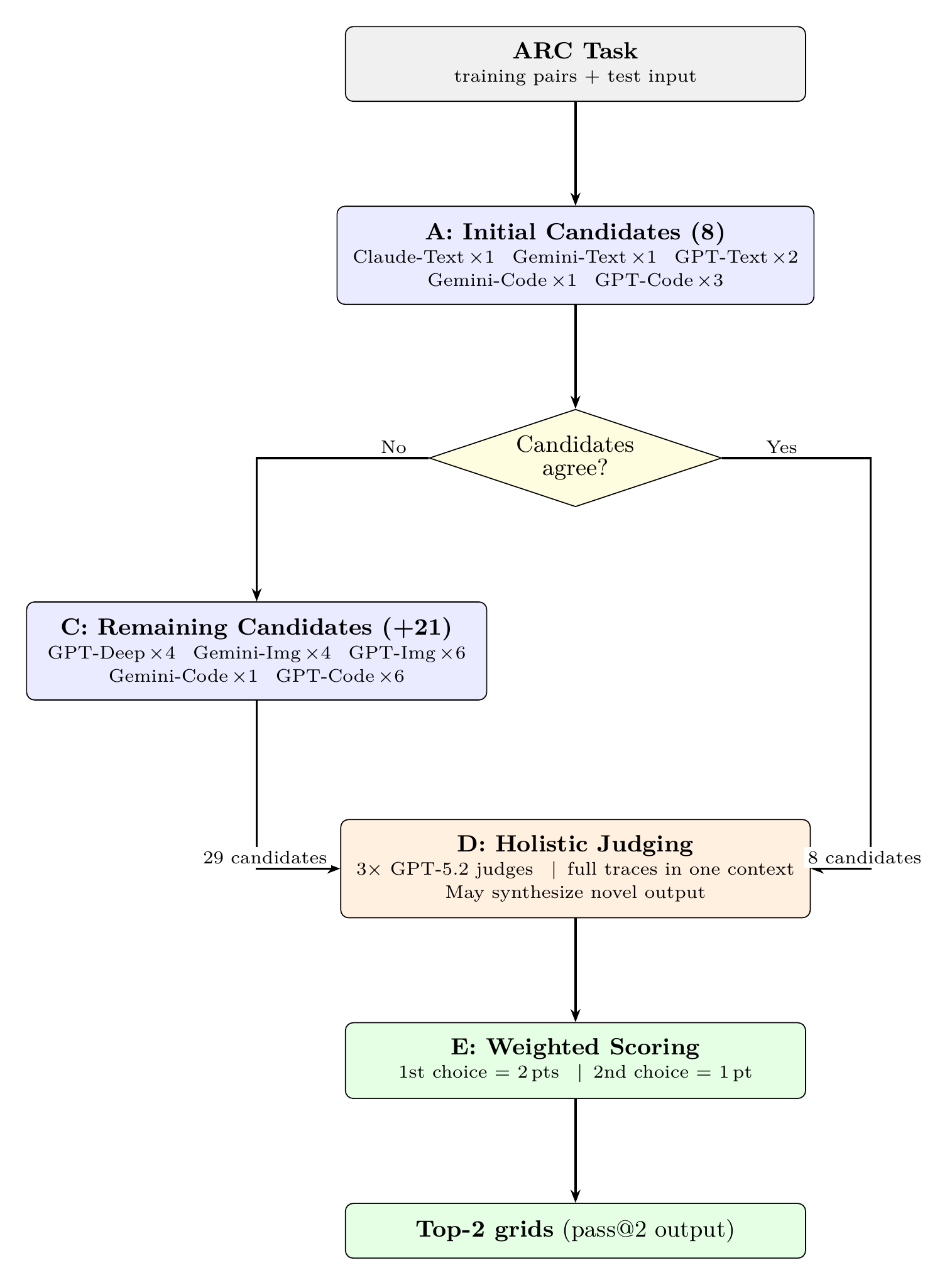}
\caption{Solver pipeline: candidate generation with adaptive early
stopping, holistic judging, and weighted scoring.}
\end{figure}

\begin{enumerate}
\def\labelenumi{\arabic{enumi}.}
\item
  \textbf{Candidate generation (up to 29 candidates).} Run a set of
  modality-specific solvers, each producing:

  \begin{itemize}
  \tightlist
  \item
    a single predicted output grid,
  \item
    a reasoning trace,
  \item
    for codegen-with-tools candidates: the code, tool calls, and tool
    outputs (execution logs).
  \end{itemize}

  Each candidate produces one output; the pass@2 two-guess format is
  introduced at the judging stage (step 2). Candidate generation
  proceeds in stages: if early-stage candidates already show strong
  agreement (multiple candidates converging on the same output), the
  system terminates early and skips the remaining, more expensive
  modalities. This adaptive early stopping improves cost efficiency on
  tasks that can be solved with a shallower search. On ARC-AGI-1 (which
  contains easier tasks that more frequently trigger early stopping),
  the same system achieves 94.5\% on the official semi-private
  evaluation at only \$11.40/task --- substantially cheaper than the
  \$38.99/task on ARC-AGI-2, where harder tasks require the full
  candidate budget more often. Both semi-private scores and costs are as
  reported by ARC Prize's verification infrastructure.\footnote{The
    semi-private evaluation is run by ARC Prize on non-public tasks; the
    author does not control that environment. Because the semi-private
    logs are not available for inspection, the detailed analysis in this
    paper (Sections 6--8) is based on the public evaluation run, where
    full data is available.}
\item
  \textbf{Holistic judging (3 parallel judges).}\\
  Concatenate all candidate traces into a single long-context prompt and
  ask a judge model to:

  \begin{itemize}
  \tightlist
  \item
    identify the top-2 most likely correct candidates (or propose a
    synthesis),
  \item
    explain why other clusters are wrong,
  \item
    output the final grids.
  \end{itemize}
\item
  \textbf{Weighted scoring.} Each judge's first choice receives 2 points
  and second choice receives 1 point. The two distinct output grids with
  the highest total score become the solver's pass@2 guesses.
\end{enumerate}

\hypertarget{development-methodology}{%
\subsection{Development methodology}\label{development-methodology}}

The solver was developed over several months during the fall of 2025,
with the author iterating on the approach, architecture, and candidate
strategies as new frontier models became available. The core
architectural ideas --- modality-driven search, independent candidate
generation, and holistic trace-based judging --- were developed and
tested against earlier model versions before the final system was
assembled with GPT-5.2 and Gemini 3 Preview in December 2025.

The codebase was built with AI-assisted development tools, and the
design process was supported by frontier AI models to explore
architectural alternatives, evaluate trade-offs between candidate
diversity strategies, and iterate on judge prompt design.

This development pattern --- using AI assistance for both design and
implementation, with the author directing strategy and evaluating
results --- proved effective for building a complex, multi-component
pipeline within a solo-researcher setting.

\hypertarget{why-candidate-diversity-matters-specifically-on-arc-agi-2}{%
\subsection{Why candidate diversity matters specifically on
ARC-AGI-2}\label{why-candidate-diversity-matters-specifically-on-arc-agi-2}}

ARC-AGI-2 tasks often introduce new concepts. A solution can be
perfectly ``logical'' on the training pairs but still be a brittle
overfit that fails to abstract the intended rule. This makes naive
``logic checking'' of traces insufficient (details in Section 5 and
Section 8). In practice, the system needs:

\begin{itemize}
\tightlist
\item
  broad hypothesis exploration across structurally different
  representations, and
\item
  a judge that can identify \textbf{where} a candidate is likely
  overfitting---even when it reads as coherent.
\end{itemize}

\begin{center}\rule{0.5\linewidth}{0.5pt}\end{center}

\hypertarget{method-multimodal-candidate-generation}{%
\section{Method: Multimodal Candidate
Generation}\label{method-multimodal-candidate-generation}}

\hypertarget{models-and-inference-configurations-used}{%
\subsection{Models and inference configurations
used}\label{models-and-inference-configurations-used}}

Candidate generation is performed via three foundation models and
multiple inference configurations:

\begin{itemize}
\tightlist
\item
  \textbf{Gemini 3 Preview}, high-reasoning setting (text and image;
  codegen with tools).
\item
  \textbf{GPT-5.2}, x-high reasoning setting (text; codegen with tools
  and without tools; image; ``deep thinking'' configuration).
\item
  \textbf{Claude Opus 4.5}, long-context (120k) (text reasoning).
\end{itemize}

\emph{(Exact API parameters, tool schemas, and prompts are documented in
the open-source implementation.)}

In the final system, candidate generators are grouped into three
families---\textbf{Text}, \textbf{Image}, and \textbf{Code}---with
multiple configurations within each family to encourage diversity. Table
1 shows the full candidate configuration.

\textbf{Table 1. Candidate configuration: 29 generators grouped by
family.}

\begin{longtable}[]{@{}lll@{}}
\toprule\noalign{}
Family & Generator & Candidates \\
\midrule\noalign{}
\endhead
\bottomrule\noalign{}
\endlastfoot
Text & Claude Opus 4.5 (text) & 1 \\
Text & Gemini 3 Preview (text) & 1 \\
Text & GPT-5.2 (text) & 2 \\
Text & GPT-5.2 (deep think) & 4 \\
Image & Gemini 3 Preview (image) & 4 \\
Image & GPT-5.2 (image) & 6 \\
Code & Gemini 3 Preview (code, tools) & 2 \\
Code & GPT-5.2 (code, tools) & 9 \\
& \textbf{Total} & \textbf{29} \\
\end{longtable}

The text family contributes 8 candidates (including 4 deep-think runs),
image contributes 10, and code contributes 11. Within each family,
multiple runs of the same generator use the same prompt and API
parameters; diversity arises from model sampling stochasticity.

\hypertarget{text-methodologies}{%
\subsection{Text methodologies}\label{text-methodologies}}

Text candidates are generated by prompting a language model with a
textual encoding of the ARC training pairs and test input. The model is
asked to infer the transformation rule and to output a completed test
grid. Text prompting is used with multiple foundation models
(Gemini/GPT/Opus) as independent candidates.

\hypertarget{base-text-prompt}{%
\subsubsection{Base text prompt}\label{base-text-prompt}}

The base prompt is intentionally minimal:

\begin{Shaded}
\begin{Highlighting}[]
\NormalTok{You are solving an ARC (Abstraction and Reasoning Corpus)}
\NormalTok{task. Each grid cell is an integer 0{-}9 representing a color.}
\NormalTok{Use the solved examples to infer the transformation and}
\NormalTok{apply it to the test input.}
\NormalTok{...}
\NormalTok{\{training and test examples\}}
\NormalTok{...}
\NormalTok{Respond with an explanation of your thinking that is detailed}
\NormalTok{enough that someone can reconstruct your solution. Afterwards,}
\NormalTok{you MUST also respond with the completed output grid.}
\end{Highlighting}
\end{Shaded}

The prompt deliberately does \textbf{not} prescribe a fixed reasoning
template, a step-by-step plan, or a fixed output grid format. This
reduces ``prompt compliance'' overhead and empirically increases
hypothesis diversity. The trade-off is that outputs are noisier and
require tolerant parsing and validation to recover candidate grids (see
also Section 8 for supporting negative results on strict output
constraints and prescribed reasoning templates).

Grids are encoded in \textbf{CSV format}, which was selected after
benchmarking 9 representation formats (standard space-separated,
semicolon-delimited, XML-tagged, CSV, Python lists, sparse coordinate
notation, ASCII symbols, binary masks, and compact pipe-delimited).
Suboptimal format choices cost on the order of 10\% lost performance
relative to CSV, with compact formats that are difficult for LLMs to
produce (e.g., sparse coordinate notation, binary masks) performing
substantially worse.

\hypertarget{deep-think-variant}{%
\subsubsection{``Deep think'' variant}\label{deep-think-variant}}

In addition to the standard text prompt, I run a ``deep think''
configuration that allocates a larger test-time compute budget. In
practice, this is achieved via a prompt modification that explicitly
encourages GPT-5.2 to reason more extensively before committing to a
final output --- effectively trading tokens for deeper deliberation
within a single response.

\hypertarget{image-methodologies}{%
\subsection{Image methodologies}\label{image-methodologies}}

Image candidates are generated by rendering the ARC training pairs and
test input as a single annotated image and providing it alongside the
(otherwise similar) instruction prompt. This provides the model with a
pixel-space representation that can be advantageous for tasks where the
salient structure is more readily perceived visually than through a
textual grid encoding.

Notably, the image renderings are intentionally \textbf{imprecise} ---
slightly distorted rather than pixel-perfect grid reproductions.
Pixel-perfect renderings underperformed in early experiments, possibly
because models treated them as lossless encodings and fell back on
cell-by-cell numerical reasoning rather than engaging the visual pattern
recognition that makes image prompting valuable in the first place. The
slight imprecision appears to encourage models to reason about shapes,
symmetries, and spatial relationships at a higher level of abstraction.

Figure 3 shows an example rendering for task \texttt{d35bdbdc:1} (public
evaluation split), where each training pair is shown as an input/output
image pair.

\begin{figure}
\centering
\includegraphics{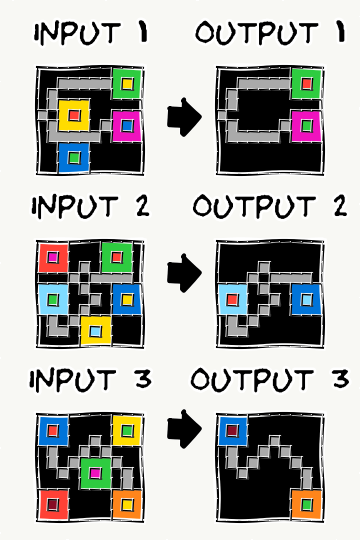}
\caption{Example image rendering used for image-based prompting (task
d35bdbdc:1).}
\end{figure}

In the public evaluation analysis (Section 6.6), image prompting
provides uniquely correct candidates on several instances, including:
\texttt{20a9e565:1}, \texttt{2d0172a1:1}, \texttt{4e34c42c:1},
\texttt{b6f77b65:1}, \texttt{b6f77b65:2}, and \texttt{d35bdbdc:1}.

\hypertarget{code-methodologies}{%
\subsection{Code methodologies}\label{code-methodologies}}

Code candidates treat ARC solving as program synthesis: the model is
asked to produce executable code that maps input grids to output grids.
I use two code-generation regimes:

\begin{enumerate}
\def\labelenumi{\arabic{enumi}.}
\tightlist
\item
  \textbf{Tool-integrated code generation (native tool calls).} The
  model iteratively writes code, executes it via a sandbox tool,
  inspects intermediate outputs, and refines the program across multiple
  tool calls.\footnote{The ARC Prize semi-private evaluation uses
    OpenAI's zero-data-retention (ZDR) API mode, which disables tool
    calls. For the semi-private run, tool-integrated code candidates
    were replaced with one-shot code generation, reducing the iterative
    refinement available to code candidates on that evaluation.}
\item
  \textbf{One-shot code generation (no tools).} The model returns a
  complete program in a single response; the harness executes it in a
  sandbox and records the result, but the model does not receive
  iterative execution feedback.
\end{enumerate}

The tool-integrated regime often produces rich intermediate artifacts
(program drafts, test harnesses, and execution traces) that are later
consumed by the holistic judge. Task \texttt{13e47133:1} is a
representative example where code-based reasoning dominates (Section
6.6). A shortened excerpt of the tool-integrated trace illustrates the
iterative development pattern:

\begin{Shaded}
\begin{Highlighting}[]
\NormalTok{"detailed\_logs": [}
\NormalTok{  \{}
\NormalTok{    "type": "code",}
\NormalTok{    "code": "import numpy as np, collections,}
\NormalTok{             itertools, math, sys, statistics\textbackslash{}n",}
\NormalTok{    "language": "python"}
\NormalTok{  \},}
\NormalTok{  ...}
\NormalTok{  \{}
\NormalTok{    "type": "code",}
\NormalTok{    "code": "\# Display expected outer right region}
\NormalTok{             \# values for rows0{-}19 cols9{-}19}
\NormalTok{             outer\_expected = ex1\_out[:,9:20]}
\NormalTok{             outer\_pred = pred1[:,9:20]}
\NormalTok{             outer\_in = ex1\_in[:,9:20]}
\NormalTok{             outer\_expected.shape\textbackslash{}n",}
\NormalTok{    "language": "python"}
\NormalTok{  \},}
\NormalTok{  ...}
\NormalTok{  \{}
\NormalTok{    "type": "code",}
\NormalTok{    "code": "\# Let\textquotesingle{}s craft final version and re{-}test.}
\NormalTok{             import numpy as np}
\NormalTok{             from collections import Counter, deque}

\NormalTok{             def solver(input\_grid):}
\NormalTok{                 g = np ...\textbackslash{}n",}
\NormalTok{    "language": "python"}
\NormalTok{  \},}
\NormalTok{  \{}
\NormalTok{    "type": "text",}
\NormalTok{    "content": "\#\#\# FINAL SOLUTION \#\#\#}
\NormalTok{               import numpy as np}
\NormalTok{               from collections import Counter, deque}

\NormalTok{               def solver(input\_grid):}
\NormalTok{                   g = np.array(input\_grid, ...)}
\NormalTok{                   h, w = g ..."}
\NormalTok{  \}}
\NormalTok{]}
\end{Highlighting}
\end{Shaded}

The one-shot (no-tools) regime is substantially cheaper and can still
contribute unique solutions, but it is less robust on tasks that benefit
from iterative debugging.

\begin{center}\rule{0.5\linewidth}{0.5pt}\end{center}

\hypertarget{method-context-preserving-holistic-judging}{%
\section{Method: Context-Preserving Holistic
Judging}\label{method-context-preserving-holistic-judging}}

\hypertarget{the-judging-problem-selecting-the-right-needle-in-a-noisy-haystack}{%
\subsection{The judging problem: selecting the right needle in a noisy
haystack}\label{the-judging-problem-selecting-the-right-needle-in-a-noisy-haystack}}

Given 8--29 candidates, many are plausible and internally coherent.
Worse, models tend to \textbf{cluster} around the same wrong
interpretation on the hardest tasks. The key difficulty is identifying a
rare candidate that is correct---or close enough that it can be
repaired---without reducing everything to an overly lossy score.

\hypertarget{judges-attempted-and-why-the-holistic-judge-wins}{%
\subsection{Judges attempted (and why the ``holistic'' judge
wins)}\label{judges-attempted-and-why-the-holistic-judge-wins}}

I tested three approaches:

\begin{enumerate}
\def\labelenumi{\arabic{enumi}.}
\item
  \textbf{Logic judge (failed mode):} score candidates by whether their
  reasoning appears logically consistent.\\
  Failure mode: a candidate can be ``logical'' yet overfit to training
  pairs or miss a newly introduced latent concept in the test case.
  Therefore logic alone is not robust.
\item
  \textbf{Consistency judge (partial):} look for themes that repeat
  across candidates.\\
  Failure mode: consistency tends to reward the majority cluster. But
  breaking new ground requires elevating \emph{divergent} hypotheses,
  because the correct solution to an unsolved task is often not the
  modal answer.
\item
  \textbf{Holistic judge (final):} provide \emph{all traces together}
  and ask the judge to pick the top-2 most likely correct candidates.
  Run three judges in parallel and aggregate via weighted scoring (see
  below). This works because \textbf{having all context together beats
  abstracting traces into scores}. It lets the judge detect subtle but
  decisive differences between near-identical hypotheses.
\end{enumerate}

The holistic judge can be understood as an \textbf{anti-consistency}
mechanism. Self-consistency (Wang et al. 2023) selects the most common
answer across samples --- an effective strategy when the majority is
likely correct. On ARC-AGI-2, the majority is often \emph{wrong} on the
hardest tasks (Section 6.8), making consistency-based selection a
liability. The holistic judge inverts this: it is designed to identify a
correct \emph{minority} hypothesis against a confidently wrong majority,
using full-trace comparison to distinguish genuine insight from
plausible groupthink.

All three judges use \textbf{GPT-5.2} (x-high reasoning setting), the
same model used for candidate generation but in a distinct role with a
different prompt. I also tested a mixed-model judge ensemble (combining
Opus, Gemini, and GPT-5.2), but three homogeneous GPT-5.2 judges
outperformed the mixed configuration. Using the same model family for
both generation and judging introduces a potential correlation risk (the
judge may favor candidates ``in its own style''), but in practice this
was outweighed by GPT-5.2's stronger individual judging capability.

\hypertarget{judge-prompt-structure}{%
\subsection{Judge prompt structure}\label{judge-prompt-structure}}

The holistic judge prompt is assembled programmatically and follows this
structure (condensed; the full implementation is in the open-source
release):

\begin{Shaded}
\begin{Highlighting}[]
\NormalTok{Below is a problem that was attempted to be solved \{N\} times:}

\NormalTok{\{training pairs + test input\}}

\NormalTok{Solutions were generated \{N\} times, using different types of solvers.}

\NormalTok{\textless{}SOLUTION 1 START\textgreater{}}
\NormalTok{\textless{}CONTENT\textgreater{}}
\NormalTok{\{full reasoning trace or extracted solver function\}}
\NormalTok{\textless{}/CONTENT\textgreater{}}
\NormalTok{\textless{}PREDICTED\_GRID\textgreater{}}
\NormalTok{\{candidate output as CSV\}}
\NormalTok{\textless{}/PREDICTED\_GRID\textgreater{}}
\NormalTok{\textless{}SOLUTION 1 STOP\textgreater{}}

\NormalTok{... (repeated for all N solutions) ...}

\NormalTok{Your task is to understand these solutions, and assess how well they\textquotesingle{}ve}
\NormalTok{understood the problem, and how likely their solutions are to provide the}
\NormalTok{correct solution to the test input.}

\NormalTok{Often, new mechanics are introduced in the test example for which the}
\NormalTok{solutions do not generalize well. Please output two solutions that you}
\NormalTok{think represent the right mechanic for solving the problem.}

\NormalTok{Output your two solutions as grids (in code blocks). Explain how you}
\NormalTok{came to these two solutions being the two most likely. Study all the}
\NormalTok{provided solutions and their reasoning to come up with a meta{-}conclusion}
\NormalTok{about how to solve the problem.}
\end{Highlighting}
\end{Shaded}

Each candidate's \texttt{CONTENT} block contains the full reasoning
trace --- for text/image candidates this is the model's chain-of-thought
response, and for code candidates it is the complete iterative tool-use
trace including intermediate program drafts, execution outputs, and
debugging steps. The \texttt{PREDICTED\_GRID} block contains the
candidate's output grid in CSV format. Candidates that produce identical
output grids are listed as separate solutions (with separate traces),
preserving the judge's ability to assess reasoning quality even when
outputs agree. The prompt does \textbf{not} instruct the judge to prefer
majority or minority answers --- it asks for a ``meta-conclusion,''
leaving the judge free to weigh agreement, reasoning quality, and
novelty as it sees fit.

\hypertarget{allowing-synthesis-new-solutions-not-in-candidates}{%
\subsection{Allowing synthesis (new solutions not in
candidates)}\label{allowing-synthesis-new-solutions-not-in-candidates}}

The holistic judge is also permitted to propose a \textbf{novel solution
not identical to any candidate output}, effectively recombining correct
subcomponents of multiple flawed candidates. This matters on tasks where
no single candidate ``gets it,'' but multiple candidates contain partial
truths. When synthesizing, the judge outputs a raw output grid directly
(not executable code), which means synthesized solutions do not benefit
from programmatic verification and are susceptible to arithmetic or
grid-construction errors.

\hypertarget{aggregation-from-3-judges-to-2-final-guesses}{%
\subsection{Aggregation: from 3 judges to 2 final
guesses}\label{aggregation-from-3-judges-to-2-final-guesses}}

Each judge outputs a ranked pair of solutions (first choice and second
choice). To produce the final pass@2 output, the system assigns
\textbf{2 points} to each judge's first choice and \textbf{1 point} to
each judge's second choice, then sums points across judges for each
distinct output grid. The two grids with the highest total score become
the solver's two guesses. When judges agree on their top pick, that
solution accumulates up to 6 points; when they disagree, the scoring
naturally surfaces the most broadly supported candidates. In practice,
full three-way disagreement on the first choice is rare on easier tasks
where candidates converge, but becomes common on harder tasks where the
judges face the same ambiguity as the generators.

\hypertarget{known-limitation-judge-biases}{%
\subsection{Known limitation: judge
biases}\label{known-limitation-judge-biases}}

Candidate traces are concatenated into the judge prompt in a fixed
order; the order is \textbf{not shuffled} between judge runs. LLMs
exhibit known position biases (e.g., favoring candidates near the start
or end of the context), and this ordering could systematically advantage
or disadvantage certain candidates. Shuffling the candidate order across
the three judge runs and measuring the effect on agreement and accuracy
is a natural improvement.

Beyond position bias, the judge may also exhibit \textbf{verbosity bias}
(favoring longer, more detailed reasoning traces over terse but correct
ones) and \textbf{format bias} (favoring candidates whose output format
more closely matches the judge's own generation patterns). These biases
are well-documented in LLM-as-a-judge settings (Zheng et al. 2023) and
could interact with the modality mix, since code candidates tend to
produce structured traces while text candidates produce prose. No
debiasing is applied in the current implementation.

\hypertarget{feasibility-and-context-length}{%
\subsection{Feasibility and context
length}\label{feasibility-and-context-length}}

The holistic judging prompt is intentionally large: on the order of
\textbf{30k--80k input tokens}, because it includes full traces from
many candidates.

This makes long-context frontier models a practical requirement for the
judge step.

\begin{center}\rule{0.5\linewidth}{0.5pt}\end{center}

\hypertarget{experiments-and-main-results}{%
\section{Experiments and Main
Results}\label{experiments-and-main-results}}

\hypertarget{evaluation-datasets-and-protocol}{%
\subsection{Evaluation datasets and
protocol}\label{evaluation-datasets-and-protocol}}

ARC-AGI-2 provides multiple evaluation sets (public, semi-private,
private), all calibrated and evaluated under \textbf{pass@2} (Chollet et
al. 2024).

ARC Prize Verified results are reported on the \textbf{semi-private
evaluation set} via an official verification process and leaderboard.

\textbf{Development data disclosure:} The solver was iteratively
designed using both the 1,000-task training set and the 120-task public
evaluation set; final configuration tuning (modality mix, candidate
counts, judge settings) was performed against the public evaluation
split. The semi-private evaluation was run on held-out tasks unseen
during development, and the resulting \textasciitilde3 percentage-point
gap (76.11\% public vs 72.9\% semi-private) suggests limited overfitting
to the public split. Additionally, the same system was run on
ARC-AGI-1's semi-private evaluation set (achieving 94.5\%\footnote{ARC-AGI-1
  result verified via the ARC Prize evaluation infrastructure:
  https://arcprize.org/leaderboard}) with \textbf{no exposure} to
ARC-AGI-1 tasks during design --- a fully blind evaluation that further
validates generalization.

\hypertarget{metrics}{%
\subsection{Metrics}\label{metrics}}

I report:

\begin{itemize}
\tightlist
\item
  \textbf{Accuracy (pass@2)}: the mean per-task solve rate, where each
  task's solve rate is the fraction of its test instances answered
  correctly within two guesses. For tasks with a single test instance
  this is binary (0 or 1); for multi-instance tasks it can be fractional
  (e.g., 0.5 if one of two test instances is solved). This is distinct
  from the \emph{instance-level} accuracy (fraction of individual test
  instances solved), which is reported separately in the per-instance
  analyses below.
\item
  \textbf{Cost per task (\$/task)}: total runtime API cost divided by
  number of tasks (including candidate generation + judging + tool calls
  + retries as applicable).
\end{itemize}

\hypertarget{headline-results}{%
\subsection{Headline results}\label{headline-results}}

\hypertarget{timeline}{%
\subsubsection{Timeline}\label{timeline}}

The solver was submitted to the ARC Prize foundation on \textbf{December
15, 2025}. Official results were announced on \textbf{February 3, 2026}.
The leaderboard snapshot in Table 2 reflects the state at the time of
announcement; subsequent entries (discussed below) have since been
added.

\hypertarget{semi-private-evaluation-arc-prize-verified-leaderboard}{%
\subsubsection{Semi-private evaluation (ARC Prize Verified /
leaderboard)}\label{semi-private-evaluation-arc-prize-verified-leaderboard}}

My solver achieves:

\begin{itemize}
\tightlist
\item
  \textbf{72.9\% solved} on ARC-AGI-2 semi-private eval (≈73\%) at
  \textbf{\$38.99/task} --- the highest score on the ARC Prize Verified
  leaderboard at the time of writing.
\end{itemize}

For context, the next two leaderboard entries at the time of the results
announcement are:

\begin{itemize}
\tightlist
\item
  \textbf{GPT-5.2 Pro}: 54.2\% at \$15.72/task
\item
  \textbf{Gemini 3 Pro}: 54.0\% at \$30.57/task
\end{itemize}

\hypertarget{public-evaluation-self-run}{%
\subsubsection{Public evaluation
(self-run)}\label{public-evaluation-self-run}}

On the public eval set, my solver achieves:

\begin{itemize}
\tightlist
\item
  \textbf{76.11\% solved} at \textbf{\$19.69/task} (self-measured).
\end{itemize}

For the per-instance analysis in Section 6.6, the public evaluation
split contains \textbf{120 task IDs} with \textbf{167 test instances}
(75 tasks with 1 test instance, 43 with 2, and 2 with 3).

ARC Prize notes that, in principle, calibrated
public/semi-private/private eval sets should be comparable when systems
are not overfit (Chollet et al. 2024). The \textasciitilde3
percentage-point gap between public (76.11\%) and semi-private (72.9\%)
likely reflects three factors: (i) natural generalization loss to a
held-out task distribution, (ii) the semi-private verification uses
OpenAI's zero-data-retention (ZDR) API mode, which disables
function/tool calling, and (iii) the semi-private run coincided with a
period of known instability in OpenAI's API, resulting in high failure
rates and extensive retries that degraded both cost and effective
candidate coverage. In this configuration, the tool-integrated code
generation candidates (Section 4.4) were replaced with one-shot code
generation (no iterative sandbox execution). Code candidates were still
produced, but without the iterative debugging loop that makes
tool-integrated generation more robust on complex tasks. Since
tool-integrated code generation accounts for the bulk of the Code
family's cost (Table 7), the ZDR constraint both reduced accuracy and
changed the cost profile of the semi-private run relative to the public
run.

\textbf{Table 2. Leaderboard snapshot and reference systems.}\footnote{https://arcprize.org/leaderboard}
Semi-private results are as reported on the ARC Prize Verified
leaderboard at the time of the official results announcement (February
3, 2026); the public-evaluation row is self-measured on the public
evaluation split.

\begin{longtable}[]{@{}
  >{\raggedright\arraybackslash}p{(\columnwidth - 8\tabcolsep) * \real{0.2000}}
  >{\raggedright\arraybackslash}p{(\columnwidth - 8\tabcolsep) * \real{0.2000}}
  >{\raggedright\arraybackslash}p{(\columnwidth - 8\tabcolsep) * \real{0.2000}}
  >{\raggedright\arraybackslash}p{(\columnwidth - 8\tabcolsep) * \real{0.2000}}
  >{\raggedright\arraybackslash}p{(\columnwidth - 8\tabcolsep) * \real{0.2000}}@{}}
\toprule\noalign{}
\begin{minipage}[b]{\linewidth}\raggedright
AI System
\end{minipage} & \begin{minipage}[b]{\linewidth}\raggedright
Author
\end{minipage} & \begin{minipage}[b]{\linewidth}\raggedright
ARC-AGI-2
\end{minipage} & \begin{minipage}[b]{\linewidth}\raggedright
Cost/Task
\end{minipage} & \begin{minipage}[b]{\linewidth}\raggedright
Comment
\end{minipage} \\
\midrule\noalign{}
\endhead
\bottomrule\noalign{}
\endlastfoot
Human Panel & Human & 100.00\% & \$17.00 & At least two humans out of
\textasciitilde400 solved it \\
This paper & Johan Land & 72.90\% & \$38.99 & Semi-private (official) \\
This paper & Johan Land & 76.11\% & \$19.69 & Public eval \\
GPT-5.2 Pro (High) & OpenAI & 54.20\% & \$15.72 & \\
Gemini 3 Pro (Refine.) & Poetiq & 54.00\% & \$30.57 & \\
GPT-5.2 (X-High) & OpenAI & 52.90\% & \$1.90 & \\
Gemini 3 Deep Think (Preview) & Google & 45.10\% & \$77.16 & \\
GPT-5.2 (High) & OpenAI & 43.30\% & \$1.39 & \\
GPT-5.2 Pro (Medium) & OpenAI & 38.50\% & \$8.99 & \\
Opus 4.5 (Thinking, 64K) & Anthropic & 37.60\% & \$2.40 & \\
Gemini 3 Flash Preview (High) & Google & 33.60\% & \$0.23 & \\
\end{longtable}

In this snapshot, the system described in this paper achieves
substantially higher verified semi-private accuracy than the strongest
single-entry commercial baselines (72.9\% vs.~\textasciitilde54\%),
indicating that modality-driven candidate generation combined with
long-context judging can move the frontier in capability. The remaining
gap to the human panel (100.0\% at \$17/task) indicates that the
benchmark still contains substantial headroom at roughly comparable
cost. The accuracy gain over commercial baselines comes at higher cost
per task relative to the cheapest entries, reflecting the additional
test-time compute spent on multi-candidate search and downstream
adjudication. The public-evaluation result (76.11\% at \$19.69/task)
suggests that comparable accuracy can be achieved at materially lower
cost on the public split, although comparisons across public
vs.~semi-private verification regimes should be interpreted cautiously.

\hypertarget{temporal-nature-of-results}{%
\subsubsection{Temporal nature of
results}\label{temporal-nature-of-results}}

The ARC-AGI-2 leaderboard is evolving rapidly, and the results reported
here should be understood as a snapshot tied to a specific moment in
frontier model development. Foundation models are improving at a pace
where single-model performance on ARC-AGI-2 can increase substantially
between model generations --- in some cases nearly doubling from one
release to the next. New model releases after the submission deadline
(December 15, 2025) have already narrowed the gap between single-model
baselines and ensemble approaches like the one described here, at a
fraction of the cost.

This trajectory is expected to continue. As base models grow stronger,
the marginal value of any fixed ensemble architecture will shift: the
same ``diverse generation + holistic judging'' pattern applied to
stronger base models should yield higher accuracy, but the gap between
the ensemble and its best single constituent will likely narrow over
time. The contribution of this paper is therefore the
\textbf{architectural pattern} --- modality-driven search paired with
context-preserving selection --- rather than the specific accuracy
numbers, which are a product of the models available at the time of
submission.

\hypertarget{efficiency-discussion}{%
\subsection{Efficiency discussion}\label{efficiency-discussion}}

ARC-AGI-2 explicitly evaluates efficiency; ARC Prize argues cost per
task is the most directly comparable efficiency axis across humans and
AI systems (Chollet et al. 2024).

Reported cost per task on both evaluation sets:

\begin{itemize}
\tightlist
\item
  \textbf{Semi-private (official):} \$38.99/task at 72.9\%
\item
  \textbf{Public eval (self-measured):} \$19.69/task at 76.11\%
\end{itemize}

For comparison, the next-best entries on the leaderboard at the time of
writing:

\begin{itemize}
\tightlist
\item
  GPT-5.2 Pro: \textbf{\$15.72/task} at 54.2\%
\item
  Gemini 3 Pro: \textbf{\$30.57/task} at 54.0\%
\end{itemize}

The roughly 2× cost difference between the semi-private and public runs
(\$38.99 vs.~\$19.69) is likely caused by API-level unreliability. Even
on the public-eval run, only 2,216 of 14,106 GPT-5.2 API attempts
succeeded (84\% failure rate due to rate limits, timeouts, and server
errors); the semi-private run, executed on ARC Prize's verification
infrastructure, likely experienced comparable or worse failure rates,
inflating cost through retried calls that did not contribute to the
final output. The public-eval cost of \textbf{\$19.69/task} is therefore
a more representative measure of the system's actual compute
requirements.

At this cost, the solver is comparable to GPT-5.2 Pro (\$15.72/task)
while achieving a +21.9 percentage-point accuracy gain (76.11\%
vs.~54.2\%), and is both cheaper and substantially more accurate than
Gemini 3 Pro (\$30.57/task at 54.0\%). A detailed cost breakdown by
component for the public-eval run is provided in Section 7.

\hypertarget{modality-contribution-and-diversity-qualitative}{%
\subsection{Modality contribution and diversity
(qualitative)}\label{modality-contribution-and-diversity-qualitative}}

The final modality mix was selected based on two criteria:

\begin{enumerate}
\def\labelenumi{\arabic{enumi}.}
\tightlist
\item
  \textbf{Performance}: raw solve contribution.
\item
  \textbf{Diversity contribution}: uniquely solved tasks that other
  modalities fail.
\end{enumerate}

Qualitatively:

\begin{itemize}
\tightlist
\item
  GPT dominates \textbf{code generation}, with Gemini adding meaningful
  diversity; Opus codegen behaved largely like a subset and was dropped
  in the final mix.
\item
  For \textbf{image reasoning}, Gemini and GPT behaved differently and
  were complementary; Opus image reasoning behaved more like a subset.
\item
  Opus was exceptional for \textbf{end-to-end text reasoning}, being the
  sole solver for several tasks via text-only reasoning.
\end{itemize}

\hypertarget{modality-complementarity-and-uniqueness-public-eval}{%
\subsection{Modality complementarity and uniqueness (public
eval)}\label{modality-complementarity-and-uniqueness-public-eval}}

To quantify complementarity between candidate-generation methodologies,
I evaluate each candidate output against the ground-truth test target
and record a per-instance correctness matrix (Figure 4). Rows correspond
to test instances (\texttt{task\_id}, \texttt{test\_index}) and columns
correspond to individual candidate generators (model × modality ×
configuration). A cell is marked PASS if the candidate output exactly
matches the ground truth, and FAIL otherwise. Blank cells indicate that
the corresponding candidate generator was not executed for that
instance.

\begin{figure}
\centering
\includegraphics{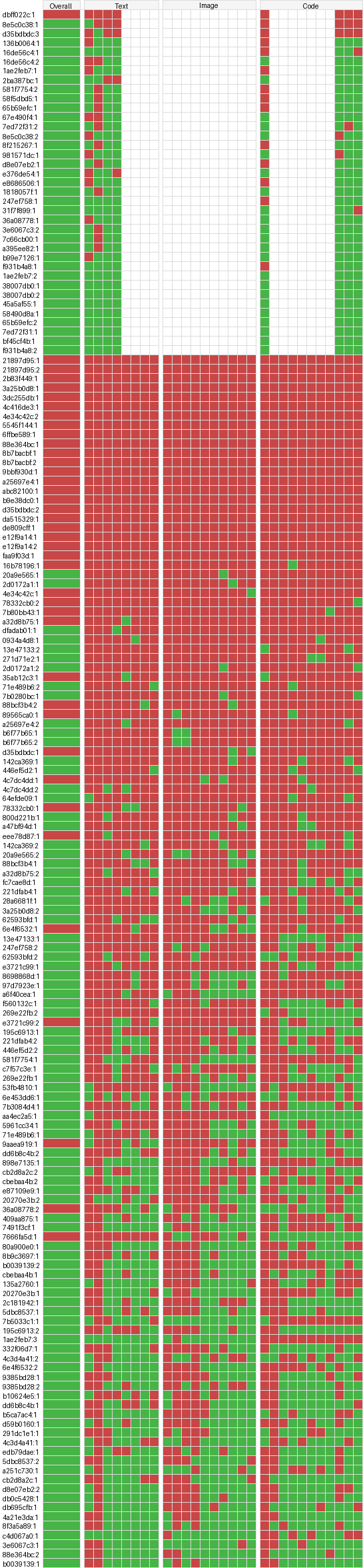}
\caption{Methodology matrix over public evaluation instances. Green =
candidate matches ground truth; red = candidate does not match; white =
candidate not produced.}
\end{figure}

Over the 167 public-evaluation test instances, the final system solves
\textbf{128/167 = 76.65\%} at the instance level (pass@2). This is
slightly higher than the task-level accuracy of 76.11\% reported in
Table 2, because partially solved multi-instance tasks pull the
task-level average below the raw instance rate. Of 120 tasks, 86 are
fully solved (all instances correct), 11 are partially solved (at least
one but not all instances correct), and 23 have no correct instances.
The candidate pool contains at least one correct output for
\textbf{144/167 = 86.23\%} of instances (candidate-oracle accuracy). The
39 unsolved instances therefore decompose into:

\begin{itemize}
\tightlist
\item
  \textbf{22/167 (13.17\%)} instances where no candidate in the pool is
  correct (candidate-generation failures).
\item
  \textbf{17/167 (10.18\%)} instances where at least one correct
  candidate exists but is not selected (selection/judging failures).
\end{itemize}

Notably, there is one instance where the final system output is correct
despite \textbf{zero} candidates matching the ground truth
(\texttt{21897d95:2}); this occurs via judge synthesis (Section 6.7),
where the judge recombines partial insights from multiple flawed
candidates to produce a novel correct output.

I operationalize modality-level uniqueness by grouping candidate
generators into three families: \textbf{Text} (including Deep),
\textbf{Image}, and \textbf{Code}. An instance is counted as solvable by
a family if any candidate within that family is correct. As described in
Section 3, the solver uses adaptive early stopping: when early-stage
candidates show strong agreement, the system skips remaining modalities
to save cost. This means 37 instances were evaluated with only 8
candidates (the initial Text + Code probe) rather than the full 29. Of
these 37 early-stopped instances, \textbf{36 were solved correctly}
(97.3\% accuracy), compared to 92/130 (70.8\%) for full-coverage
instances. Only one early-stopped instance (\texttt{dbff022c:1},
https://arcprize.org/play?task=dbff022c) was incorrectly solved --- a
case of extreme groupthink. The task itself is relatively simple, but
the test case introduces a new mechanic with two valid interpretations:
the legend that maps symbols to colors is either ``the same'' as in
training or ``inverted.'' All models confidently assume the simpler
interpretation (``legend is the same''), while the ground truth requires
the more complex one (``legend is inverted''). This is arguably an
artificial source of difficulty --- the ambiguity is not clearly
disambiguated by the training examples --- but it illustrates a
characteristic challenge of ARC: test cases can introduce subtle twists
that render the majority hypothesis wrong, and no amount of additional
candidates would help when every model makes the same simplifying
assumption. This high accuracy rate (36/37) validates the early-stopping
heuristic, but only for the easiest tasks --- which are precisely the
tasks that trigger early stopping. On harder tasks, candidates do not
converge, early stopping is not triggered, and the full 29-candidate
budget is spent. It is on these harder tasks that the system's core
contributions --- modality diversity and holistic judging --- become
essential, because majority voting systematically fails when the correct
solution is a minority hypothesis (Section 5, Section 7).

To avoid conflating uniqueness with conditional execution, I restrict
the modality-level analysis below to the \textbf{130 instances with
complete candidate coverage} (29/29 candidate columns filled).

\textbf{Table 3. Pairwise non-overlap between modality families on the
complete-coverage subset (n = 130).} Each cell reads ``row solves,
column does not'': the entry counts instances with at least one PASS in
the row family and zero PASS in the column family. The table is not
symmetric.

\begin{longtable}[]{@{}llll@{}}
\toprule\noalign{}
& Text & Image & Code \\
\midrule\noalign{}
\endhead
\bottomrule\noalign{}
\endlastfoot
Text & NA & 13 & 7 \\
Image & 17 & NA & 11 \\
Code & 18 & 18 & NA \\
\end{longtable}

\textbf{Table 4. Exclusive coverage on the complete-coverage subset (n =
130).} An instance is exclusive to a family if that family has at least
one PASS and the other two families have zero PASS.

\begin{longtable}[]{@{}ll@{}}
\toprule\noalign{}
Family & Count \\
\midrule\noalign{}
\endhead
\bottomrule\noalign{}
\endlastfoot
Text only & 2 \\
Image only & 6 \\
Code only & 7 \\
\end{longtable}

Taken together, Tables 3--4 indicate substantial complementarity between
modality families: each family covers a non-trivial set of instances
that are not covered by at least one of the other families, and
exclusive coverage persists even when all candidates are generated. This
supports treating modalities as distinct search operators rather than
relying on a single representation.

Beyond these aggregate counts, Figure 4 exhibits pronounced
\emph{instance-level heterogeneity}: some tasks are solved reliably by
one family while being largely unsolved by others. For example, task
\texttt{a6f40cea} (https://arcprize.org/play?task=a6f40cea) is solved by
\textbf{7/10} Image candidates, \textbf{1/8} Text candidates, and
\textbf{0/11} Code candidates on its test instance. Qualitatively, the
underlying rule resembles a localized magnification or ``lens''
operation (identifying a small region and ``zooming'' it), which is
naturally expressed and discovered in pixel space. Conversely, task
\texttt{13e47133} (https://arcprize.org/play?task=13e47133) is solved
\emph{only} by Code candidates on both of its test instances (Text =
0/8, Image = 0/10, Code = 7/11 and 2/11, respectively). In this case,
the solution is readily implemented by explicitly traversing the
boundary/edge structure, a representation that is directly available to
code-based generators but comparatively indirect for pure text or visual
prompt-based reasoning.

\hypertarget{judging-and-synthesis-public-eval}{%
\subsection{Judging and synthesis (public
eval)}\label{judging-and-synthesis-public-eval}}

\hypertarget{judge-based-selection-excluding-synthesis}{%
\subsubsection{Judge-based selection (excluding
synthesis)}\label{judge-based-selection-excluding-synthesis}}

Judging (excluding synthesis) had a net uplift of \textbf{7} instances
that were solved relative to a majority-vote baseline (selecting the
most common candidate output). All 7 are minority recoveries --- cases
where the correct answer was not the majority output and the holistic
judge identified it by reasoning over the full traces. One example of
this is \texttt{dfadab01:1} (https://arcprize.org/play?task=dfadab01).
This problem heavily suffers from ``group think'': 12 of the candidate
solvers converge on the same incorrect output, and another 8 converge on
a second incorrect output. Only one candidate solver produces the
correct output. The judges identify the originality in this lone
solution and select it. The judge rationale emphasizes that most
candidates recover the same underlying stamp mechanic, and that the
remaining uncertainty is restricted to edge handling:

\begin{Shaded}
\begin{Highlighting}[]
\NormalTok{Most candidate solvers correctly identify the}
\NormalTok{core stamp mechanic:}
\NormalTok{...}
\NormalTok{I prefer solutions that do *not* stamp at (9,8)}
\NormalTok{(e.g., solutions 10{-}15/17{-}20) over those that}
\NormalTok{stamp every 8.}

\NormalTok{The only remaining ambiguity is edge handling}
\NormalTok{(not clearly disambiguated by the training set):}
\NormalTok{{-} Some solvers assume a stamp must fit fully}
\NormalTok{  (ignore row 17 markers).}
\NormalTok{{-} Others assume stamps are clipped at the border}
\NormalTok{  (row 17 markers produce the top 3 rows of}
\NormalTok{  the tile).}

\NormalTok{So the two most plausible outputs}
\NormalTok{(same mechanic, differing only in border handling)}
\end{Highlighting}
\end{Shaded}

The judge correctly identifies that the core mechanic is not in dispute
--- all candidates agree on the stamp logic --- and reasons that the
real ambiguity lies in edge handling, which the training examples do not
disambiguate. Rather than committing both guesses to the majority
interpretation, the judge uses the pass@2 format to hedge: it selects
one output for each interpretation of the border behavior, ensuring that
the correct answer is captured regardless of which edge-handling rule
the ground truth expects. This is precisely the kind of reasoning that
majority voting cannot perform --- it requires understanding
\emph{where} the uncertainty lies and deliberately including a minority
hypothesis to cover it.

\hypertarget{judge-synthesis}{%
\subsubsection{Judge synthesis}\label{judge-synthesis}}

The judge is permitted to output a \textbf{synthesized} solution that is
not identical to any single candidate output (Section 5.3). On the
public evaluation run summarized in Figure 4, synthesis was invoked
\textbf{17} times total across all three judges. Relative to an ablation
that disables synthesis (i.e., forces selection among non-synthesized
candidates), synthesis yields an uplift of \textbf{1} additional solved
instance.

The single instance solved via synthesis is \texttt{21897d95:2}
(https://arcprize.org/play?task=21897d95) --- a particularly difficult
task where \textbf{none} of the 29 candidates produced a correct output
for this test instance. The task requires identifying ``rooms'' in the
grid separated by boundaries, recognizing T-shaped arrows that encode
directional recoloring instructions, and applying these instructions
across room boundaries. The judge synthesized a correct output by
recombining partial insights from multiple flawed candidates ---
identifying which candidates correctly parsed the room structure, which
correctly interpreted the arrow semantics, and which correctly handled
the recoloring --- even though no single candidate got the full chain
right. Below is an excerpt from the judge's synthesis rationale:

\begin{Shaded}
\begin{Highlighting}[]
\NormalTok{{-}{-}{-} COUNCIL OF JUDGES SYNTHESIZED SOLUTION {-}{-}{-}}
\NormalTok{Across the training examples, the only solutions that}
\NormalTok{consistently match the *actual* structure are SOLUTION 1}
\NormalTok{and SOLUTION 2 (they are the same). They correctly}
\NormalTok{identify the key mechanic:}

\NormalTok{{-} The input is divided into large "rooms" (rectangular}
\NormalTok{  regions of a dominant background color).}
\NormalTok{{-} Inside rooms, color 1 forms a 4{-}cell T{-}shape arrow}
\NormalTok{  (a 3{-}long bar + 1{-}cell stem).}
\NormalTok{{-} The stem direction indicates which neighboring room}
\NormalTok{  is the target.}
\NormalTok{{-} The paint color is:}
\NormalTok{  {-} the center of the 3{-}long bar if it is not 1}
\NormalTok{    (a payload color), otherwise}
\NormalTok{  {-} the background color of the room containing}
\NormalTok{    the arrow.}
\NormalTok{...}
\NormalTok{the same room recoloring but then rotated 90 CCW.}
\NormalTok{This is much less likely given the square training}
\NormalTok{examples, but it matches the extra rotation behavior}
\NormalTok{seen in the non{-}square examples and is the most}
\NormalTok{plausible "geometry variant" if a solver applied}
\NormalTok{that step unconditionally}
\NormalTok{...}
\end{Highlighting}
\end{Shaded}

This example illustrates the core value proposition of synthesis: on
tasks where the reasoning chain is too long for any single candidate to
get right, the judge can act as a meta-reasoner that assembles the
correct solution from complementary fragments. The current
implementation remains under-optimized; further work on synthesis gating
(Section 9) could increase the realized uplift, particularly on harder
tasks where this recombination mechanism has the most potential.

\hypertarget{failure-analysis-public-eval}{%
\subsection{Failure analysis (public
eval)}\label{failure-analysis-public-eval}}

Of the 39 unsolved test instances, 21 are \textbf{generation failures}
(no candidate in the pool is correct), 17 are \textbf{selection
failures} (at least one correct candidate exists but is not selected by
the judge), and 1 (\texttt{dbff022c:1}) is an \textbf{early-stopping
failure} discussed in Section 6.6 --- that instance was stopped after 8
candidates due to apparent consensus, so image and most code candidates
were never generated, and it cannot be classified as a generation
failure across all modalities. This section lists the generation and
selection failure groups to support qualitative analysis of what makes
these instances hard.

\hypertarget{generation-failures-21-instances}{%
\subsubsection{Generation failures (21
instances)}\label{generation-failures-21-instances}}

The following instances have zero correct candidates across all
modalities (with full 29-candidate coverage). These represent tasks
where the system's hypothesis space --- across text, image, and code ---
does not contain the ground-truth transformation.

Many of these tasks appear to require \textbf{long chains of dependent
reasoning steps}, where the correct solution emerges only after
composing multiple sub-concepts in sequence. Each step narrows the
interpretation space, but models tend to collapse or shortcut the chain
rather than faithfully executing all steps. Three illustrative examples:

\begin{itemize}
\tightlist
\item
  \textbf{\texttt{3dc255db}} (https://arcprize.org/play?task=3dc255db)
  requires recognizing ``spaceships'' with distinct inner and outer
  regions and a directional front; the outer areas are preserved while
  inner areas lengthen the front, with the inner/outer distinction
  determined by the longest edge. This involves at least four dependent
  inferences (shape identification, region segmentation, directional
  semantics, and edge-length-based classification).
\item
  \textbf{\texttt{88e364bc}} (https://arcprize.org/play?task=88e364bc)
  requires interpreting a legend that encodes movement directions, then
  simulating movement that respects borders and colors, across multiple
  independently placed legend entries and areas. The combination of
  legend parsing, spatial simulation, and multi-region generalization
  defeats all modalities.
\item
  \textbf{\texttt{d35bdbdc}} (test 2,
  https://arcprize.org/play?task=d35bdbdc) involves identifying a path
  structure, distinguishing inner from outer colors, deleting non-path
  elements, and recursively transferring colors --- with shapes that
  vary in form across instances.
\end{itemize}

These examples suggest that the system's candidate generators can
sometimes identify individual sub-concepts but struggle to compose the
full chain correctly. This is consistent with known limitations of LLM
reasoning on deeply compositional tasks, and points to a potential
benefit of multi-step decomposition approaches --- though Section 8
documents that naive decomposition strategies reduced diversity in
practice.

\begin{longtable}[]{@{}lll@{}}
\toprule\noalign{}
Task & Test & Link \\
\midrule\noalign{}
\endhead
\bottomrule\noalign{}
\endlastfoot
\texttt{21897d95} & 1 & https://arcprize.org/play?task=21897d95 \\
\texttt{2b83f449} & 1 & https://arcprize.org/play?task=2b83f449 \\
\texttt{3a25b0d8} & 1 & https://arcprize.org/play?task=3a25b0d8 \\
\texttt{3dc255db} & 1 & https://arcprize.org/play?task=3dc255db \\
\texttt{4c416de3} & 1 & https://arcprize.org/play?task=4c416de3 \\
\texttt{4e34c42c} & 2 & https://arcprize.org/play?task=4e34c42c \\
\texttt{5545f144} & 1 & https://arcprize.org/play?task=5545f144 \\
\texttt{6ffbe589} & 1 & https://arcprize.org/play?task=6ffbe589 \\
\texttt{88e364bc} & 1 & https://arcprize.org/play?task=88e364bc \\
\texttt{8b7bacbf} & 1 & https://arcprize.org/play?task=8b7bacbf \\
\texttt{8b7bacbf} & 2 & https://arcprize.org/play?task=8b7bacbf \\
\texttt{9bbf930d} & 1 & https://arcprize.org/play?task=9bbf930d \\
\texttt{a25697e4} & 1 & https://arcprize.org/play?task=a25697e4 \\
\texttt{abc82100} & 1 & https://arcprize.org/play?task=abc82100 \\
\texttt{b9e38dc0} & 1 & https://arcprize.org/play?task=b9e38dc0 \\
\texttt{d35bdbdc} & 2 & https://arcprize.org/play?task=d35bdbdc \\
\texttt{da515329} & 1 & https://arcprize.org/play?task=da515329 \\
\texttt{de809cff} & 1 & https://arcprize.org/play?task=de809cff \\
\texttt{e12f9a14} & 1 & https://arcprize.org/play?task=e12f9a14 \\
\texttt{e12f9a14} & 2 & https://arcprize.org/play?task=e12f9a14 \\
\texttt{faa9f03d} & 1 & https://arcprize.org/play?task=faa9f03d \\
\end{longtable}

\hypertarget{selection-failures-17-instances}{%
\subsubsection{Selection failures (17
instances)}\label{selection-failures-17-instances}}

The following instances have at least one correct candidate, but the
holistic judge fails to select it. These represent cases where the
judge's selection mechanism leads to an incorrect final output.

A qualitative observation across several selection failures is that the
\textbf{test instance introduces a new mechanic or configuration not
fully disambiguated by the training examples}, creating genuine
ambiguity about the correct generalization. In these cases, some
candidates make the right assumptions about how the rule extends, but
the judge --- reasoning from the same ambiguous training pairs --- tends
to favor the majority interpretation. Two illustrative examples:

\textbf{\texttt{36a08778}} (test 2,
https://arcprize.org/play?task=36a08778): The training examples
establish a straightforward ``water flows downward'' mechanic. Test 2
introduces walls that block flow --- a new structural element not
present in training. Most candidates (and the judge) converge on the
simpler mechanic, discarding the minority candidates that correctly
handle walls:

\begin{Shaded}
\begin{Highlighting}[]
\NormalTok{Most of the 29 solvers converged to the *same* mechanic (and the same output)}
\NormalTok{...}
\NormalTok{I discarded outliers like **Solution 2** and **Solution 1**}
\end{Highlighting}
\end{Shaded}

\textbf{\texttt{88bcf3b4}} (test 2,
https://arcprize.org/play?task=88bcf3b4): Training examples show a
single ``rope/snake'' component moving in one direction. Test 2
introduces multiple strings moving in multiple directions --- a
generalization the training pairs do not disambiguate. The judge
identifies the ambiguity but cannot resolve it, and selects the wrong
resolution:

\begin{Shaded}
\begin{Highlighting}[]
\NormalTok{From the 5 training examples, the consistent mechanic is **not** "gravity" or}
\NormalTok{"attraction" of whole blobs. Instead, one non{-}background component acts like a}
\NormalTok{**rope/snake**}
\NormalTok{...}
\NormalTok{The main ambiguity the examples do *not* disambiguate is what happens when the}
\NormalTok{"returning" segment reaches the pole\textquotesingle{}s column/row again:}
\NormalTok{{-} **Solution 20** continues the diagonal return even after hitting the pole\textquotesingle{}s}
\NormalTok{  column (so it can pass "through" and go beyond).}
\NormalTok{{-} **Solution 16** effectively **clamps** once aligned with the pole\textquotesingle{}s column}
\NormalTok{  (keeps going straight instead of overshooting).}
\end{Highlighting}
\end{Shaded}

This failure mode is structurally difficult for the holistic judge: when
training examples are consistent with multiple generalizations and the
test instance is the only signal that distinguishes them, the judge
faces the same underspecification that makes the task hard in the first
place. The judge's tendency to favor majority-cluster agreement ---
which is beneficial on tasks where the majority is correct (e.g.,
\texttt{dfadab01:1} in Section 6.7) --- becomes a liability on tasks
where the correct generalization is a minority hypothesis precisely
because it requires handling a novel test-time mechanic.

\begin{longtable}[]{@{}lll@{}}
\toprule\noalign{}
Task & Test & Link \\
\midrule\noalign{}
\endhead
\bottomrule\noalign{}
\endlastfoot
\texttt{16b78196} & 1 & https://arcprize.org/play?task=16b78196 \\
\texttt{35ab12c3} & 1 & https://arcprize.org/play?task=35ab12c3 \\
\texttt{36a08778} & 2 & https://arcprize.org/play?task=36a08778 \\
\texttt{4c7dc4dd} & 1 & https://arcprize.org/play?task=4c7dc4dd \\
\texttt{4e34c42c} & 1 & https://arcprize.org/play?task=4e34c42c \\
\texttt{6e4f6532} & 1 & https://arcprize.org/play?task=6e4f6532 \\
\texttt{7666fa5d} & 1 & https://arcprize.org/play?task=7666fa5d \\
\texttt{78332cb0} & 1 & https://arcprize.org/play?task=78332cb0 \\
\texttt{78332cb0} & 2 & https://arcprize.org/play?task=78332cb0 \\
\texttt{7b80bb43} & 1 & https://arcprize.org/play?task=7b80bb43 \\
\texttt{88bcf3b4} & 2 & https://arcprize.org/play?task=88bcf3b4 \\
\texttt{89565ca0} & 1 & https://arcprize.org/play?task=89565ca0 \\
\texttt{9aaea919} & 1 & https://arcprize.org/play?task=9aaea919 \\
\texttt{a32d8b75} & 1 & https://arcprize.org/play?task=a32d8b75 \\
\texttt{d35bdbdc} & 1 & https://arcprize.org/play?task=d35bdbdc \\
\texttt{e3721c99} & 2 & https://arcprize.org/play?task=e3721c99 \\
\texttt{eee78d87} & 1 & https://arcprize.org/play?task=eee78d87 \\
\end{longtable}

\begin{center}\rule{0.5\linewidth}{0.5pt}\end{center}

\hypertarget{ablation-studies}{%
\section{Ablation Studies}\label{ablation-studies}}

A rigorous component attribution would require multiple full-pipeline
runs with individual components removed or substituted. Each such run
costs approximately \$2,400 in API spend (Table 6), making extensive
ablation prohibitively expensive --- a full ablation matrix across
modalities, judge configurations, and candidate budgets would cost tens
of thousands of dollars. This paper therefore relies primarily on
\textbf{post-hoc analysis of the single public evaluation run},
extracting what can be measured from the existing data (e.g., comparing
judge selections against majority-vote baselines on the same candidate
pool) rather than running dedicated ablation experiments. This approach
has clear limitations: it cannot capture interaction effects between
components, and some comparisons (e.g., end-to-end modality removal)
require fresh runs that have not been performed. The ablations reported
here should be read with this constraint in mind.

Unless otherwise stated, ``solved'' refers to pass@2 at the
\textbf{test-instance} level on the public evaluation split (167
instances).

\hypertarget{measured-ablations}{%
\subsection{Measured ablations}\label{measured-ablations}}

\hypertarget{judging-holistic-selection-vs-per-candidate-scoring-excluding-synthesis}{%
\subsubsection{Judging: holistic selection vs per-candidate scoring
(excluding
synthesis)}\label{judging-holistic-selection-vs-per-candidate-scoring-excluding-synthesis}}

Judging (excluding synthesis) had a net uplift of \textbf{7} solved
instances relative to a \textbf{majority-vote baseline} that selects the
most common candidate output grid as the first guess and the
second-most-common as the second guess (i.e., standard
self-consistency). All 7 uplift instances are \textbf{minority
recoveries}: cases where the correct answer was not the most frequent
candidate output, and the holistic judge identified it by reasoning over
the full traces rather than counting votes. This directly validates the
``anti-consistency'' motivation (Section 5): on these tasks, the
majority cluster was wrong, and the holistic judge's ability to read and
compare reasoning traces --- rather than simply tallying outputs --- was
the deciding factor. One illustrative instance is \texttt{dfadab01:1}
(https://arcprize.org/play?task=dfadab01), where the candidate pool
clusters around two distinct incorrect hypotheses and the holistic judge
selects the lone correct candidate.

\hypertarget{judging-synthesis-enabled-vs-disabled}{%
\subsubsection{Judging: synthesis enabled vs
disabled}\label{judging-synthesis-enabled-vs-disabled}}

Synthesized solution yields a net uplift of \textbf{1} additional solved
instance in this run. Across all three judges, synthesis was invoked
\textbf{17} times total; in most cases the synthesized output did not
change the final selected solution, as the weighted scoring still
favored non-synthesized candidates.

While the measured uplift is small on this particular run, synthesis has
been a more material contributor in other experimental runs during
development. The mechanism has particular potential on harder tasks
where no single candidate is fully correct but multiple candidates
contain complementary partial insights --- exactly the regime where
recombination should help most. Better understanding when and how to
trigger synthesis (see ``synthesis gating'' in Section 9) is an
important direction for future work.

\textbf{Table 5. Measured judge ablations on the public evaluation run.}
Reported deltas are net solved-instance uplifts attributable to the
indicated component (with the stated control condition).

\begin{longtable}[]{@{}
  >{\raggedright\arraybackslash}p{(\columnwidth - 4\tabcolsep) * \real{0.3333}}
  >{\raggedright\arraybackslash}p{(\columnwidth - 4\tabcolsep) * \real{0.3333}}
  >{\raggedright\arraybackslash}p{(\columnwidth - 4\tabcolsep) * \real{0.3333}}@{}}
\toprule\noalign{}
\begin{minipage}[b]{\linewidth}\raggedright
Component
\end{minipage} & \begin{minipage}[b]{\linewidth}\raggedright
Ablation / control condition
\end{minipage} & \begin{minipage}[b]{\linewidth}\raggedright
Net uplift (solved instances)
\end{minipage} \\
\midrule\noalign{}
\endhead
\bottomrule\noalign{}
\endlastfoot
Holistic selection & Holistic judge vs majority-vote baseline (synthesis
disabled in both settings) & +7 (all minority recoveries) \\
Judge synthesis & Synthesis enabled vs disabled (holistic selection held
fixed) & +1 \\
\end{longtable}

\hypertarget{cost-attribution-per-component}{%
\subsubsection{Cost attribution per
component}\label{cost-attribution-per-component}}

Table 6 reports the cost breakdown for the public evaluation run,
averaged per test instance (i.e., per task--test-pair, of which there
are 167). This is distinct from the per-task cost reported in Table 2
(\$19.69/task over 120 tasks), because many tasks contain multiple test
instances (75 tasks have 1 test instance, 43 have 2, and 2 have 3). The
per-instance average (\$14.31) does not sum to the per-task figure
because the per-task metric aggregates all test instances within a task
into a single cost. Note: a strict roll-up of the Table 6 total
(\$2,390.28 / 120 tasks = \$19.92/task) differs slightly from the
reported \$19.69/task; the discrepancy is likely due to accumulated
floating-point rounding in the cost-accounting script.

\textbf{Table 6. Cost attribution per test instance on the public
evaluation run (n = 167).}

\begin{longtable}[]{@{}llll@{}}
\toprule\noalign{}
Phase & Total (\$) & Avg \$/instance & \% of total \\
\midrule\noalign{}
\endhead
\bottomrule\noalign{}
\endlastfoot
Candidate generation & 2081.37 & 12.46 & 87.1\% \\
Judging & 308.91 & 1.85 & 12.9\% \\
\textbf{Total} & \textbf{2390.28} & \textbf{14.31} & \textbf{100\%} \\
\end{longtable}

\textbf{Table 7. Candidate generation cost by modality family.}

\begin{longtable}[]{@{}llll@{}}
\toprule\noalign{}
Modality family & Total (\$) & Avg \$/instance & \% of generation
cost \\
\midrule\noalign{}
\endhead
\bottomrule\noalign{}
\endlastfoot
Text (incl.~deep think) & 597.70 & 3.58 & 28.7\% \\
Image & 467.10 & 2.80 & 22.5\% \\
Code & 1016.56 & 6.09 & 48.9\% \\
\end{longtable}

Candidate generation dominates overall cost at 87\% of spend, with
judging accounting for only 13\%. Within generation, code candidates are
the most expensive family (49\% of generation cost), driven primarily by
tool-integrated code generation with iterative sandbox execution. Given
that judging contributes +7 solved instances (holistic selection) and +1
(synthesis) at only 13\% of total cost, the judging phase is highly
cost-effective relative to its accuracy contribution.

\hypertarget{modality-ablations-oracle-level-only}{%
\subsection{Modality ablations (oracle-level
only)}\label{modality-ablations-oracle-level-only}}

Section 6.6 reports modality-level uniqueness on the
\textbf{complete-coverage subset} (n = 130), where all modalities are
executed. In this subset, the following \textbf{exclusive} oracle
solvability counts are observed (Table 4): Text only = 2, Image only =
6, Code only = 7. These exclusive counts imply that removing any single
modality would reduce candidate-oracle coverage by at least a few
percent even before accounting for downstream judge interactions and
selection effects.

However, oracle-level analysis has an important limitation: it measures
whether a correct candidate \emph{exists} in a modality's output, but
does not measure the end-to-end effect of removing that modality on the
final system output. Removing a modality could affect judge behavior in
ways not captured by oracle overlap --- for instance, reducing the
number of candidates changes cluster dynamics, which could make it
easier or harder for the holistic judge to identify the correct
solution. A modality might also contribute ``near-miss'' candidates that
inform judge synthesis even when no individual candidate from that
modality is exactly correct.

The proper ablation --- running the full pipeline with one modality
family removed and re-running judging on the reduced candidate pool ---
has not been performed. Each such run requires regenerating all judge
transcripts on the reduced candidate set (and ideally multiple
repetitions to account for judge variance), making it expensive relative
to the oracle-level analysis. This remains a gap; the oracle-level
uniqueness numbers in Section 6.5 should be interpreted as a lower bound
on each modality's contribution, not as a precise end-to-end
attribution.

\hypertarget{unperformed-ablations}{%
\subsection{Unperformed ablations}\label{unperformed-ablations}}

The following ablations would strengthen the paper's claims but have not
been run due to cost constraints. They are listed here both as
transparency about what remains unknown and as a roadmap for future
work.

\textbf{Generation ablations:}

\begin{itemize}
\tightlist
\item
  \textbf{End-to-end modality removal:} run the full pipeline with one
  modality family (text, image, or code) removed and re-run judging on
  the reduced candidate pool. The oracle-level exclusive counts in
  Section 7.2 provide a lower bound, but the actual end-to-end impact
  --- including judge interaction effects --- is unknown. This requires
  at minimum three full runs (\textasciitilde\$7,200 total).
\item
  \textbf{Independent candidates vs sequential refinement:} hold compute
  fixed and compare N independent candidates (the current approach)
  against N sequential refinement steps (iterative prompt chaining or
  staged decomposition). Section 8 documents qualitative evidence that
  sequential approaches reduced diversity, but a controlled comparison
  with matched compute budgets has not been performed.
\item
  \textbf{Candidate budget scaling:} sweep the number of candidates per
  modality/model to estimate marginal returns per additional candidate
  and identify diminishing-returns regimes. The current system uses 29
  candidates, but it is unknown whether 15 or 50 would yield
  meaningfully different accuracy at proportionally different cost.
\item
  \textbf{Per-model contribution:} isolate the contribution of each
  foundation model (GPT-5.2, Gemini 3, Opus 4.5) by running the pipeline
  with one model removed entirely. Opus 4.5 contributes only 1
  candidate; whether its inclusion is cost-justified relative to adding
  another GPT-5.2 or Gemini candidate is unknown.
\item
  \textbf{Temperature and sampling parameters:} the current system uses
  default or near-default sampling settings for each model. Sweeping
  temperature, top-p, and other sampling parameters within each modality
  could reveal whether diversity is better increased through sampling
  variation or through modality variation.
\item
  \textbf{Representation formats:} CSV vs alternative encodings (e.g.,
  JSON-like arrays, Python list syntax, and object-abstraction
  encodings), evaluated under the same candidate/judge budgets. The
  benchmarking reported in Section 4 was performed during development
  with different model versions; a controlled evaluation on the final
  system would be more rigorous.
\end{itemize}

\textbf{Selection and judging ablations:}

\begin{itemize}
\tightlist
\item
  \textbf{Full majority-vote baseline comparison:} the +7 holistic
  selection uplift reported above is computed post hoc by comparing
  judge selections against the majority output grid on the same
  candidate pool. A cleaner comparison would run majority vote as the
  \emph{sole} selection mechanism in a full end-to-end run (without any
  judge invocation), eliminating any confounds from shared
  infrastructure. This would also quantify whether the 13\% cost of
  judging is justified by the accuracy gain.
\item
  \textbf{Judge ensemble sizing:} the current system uses 3 judges with
  weighted scoring. No ablation comparing 1-judge vs 3-judge accuracy
  has been performed. Quantifying the disagreement rate and how often
  the ensemble corrects vs overrides individual judges would clarify
  whether the ensemble cost is justified or whether a single judge
  suffices.
\item
  \textbf{Alternative selection mechanisms:} compare the holistic judge
  against other selection strategies on the same candidate pool,
  including per-output log-probability scoring, pairwise judge
  tournaments (comparing candidates two at a time), and best-of-N with a
  reward model. These comparisons would isolate the contribution of
  joint-context evaluation from other factors.
\item
  \textbf{Judge model diversity:} the current system uses three GPT-5.2
  judges. A mixed-model ensemble (e.g., one GPT-5.2, one Gemini, one
  Opus judge) was tested informally and underperformed (Section 5), but
  a rigorous comparison --- controlling for prompt format and scoring
  calibration --- has not been done.
\item
  \textbf{Trace content ablation:} the holistic judge receives full
  reasoning traces alongside candidate outputs. Comparing judge accuracy
  with traces vs outputs-only would quantify whether the trace content
  actually helps selection or whether the judge primarily relies on
  output grid comparison.
\end{itemize}

\textbf{Early stopping ablations:}

\begin{itemize}
\tightlist
\item
  \textbf{Early stopping threshold tuning:} the current heuristic
  triggers early stopping when initial candidates agree. Varying the
  agreement threshold and the number of candidates consulted before the
  stopping decision would characterize the accuracy/cost trade-off and
  the groupthink risk documented in Section 6.6.
\end{itemize}

\begin{center}\rule{0.5\linewidth}{0.5pt}\end{center}

\hypertarget{negative-results-and-discarded-approaches}{%
\section{Negative Results and Discarded
Approaches}\label{negative-results-and-discarded-approaches}}

This section documents explored approaches that were ultimately
discarded, often because they reduced diversity, increased brittleness,
or forced premature abstraction.

\hypertarget{hint-generation-solver-discarded}{%
\subsection{Hint generation → solver
(discarded)}\label{hint-generation-solver-discarded}}

This approach is structurally similar to iterative self-improvement
methods such as Self-Refine (Madaan et al. 2023) and Reflexion (Shinn et
al. 2023), where an initial pass generates feedback that informs a
subsequent attempt.

Motivation: - doubling the reasoning budget across two turns (hint then
solve).

Observed drawback: - the hint stage often \textbf{limits creativity} and
collapses candidate diversity into a narrower space, which is
counterproductive when trying to break new ground.

\hypertarget{object-identification-transformation-identification-solver-discarded}{%
\subsection{Object identification → transformation identification →
solver
(discarded)}\label{object-identification-transformation-identification-solver-discarded}}

Motivation: - structured decomposition to ``force'' abstraction.

Failure mode: - brittle handoff between stages. Both verbose and overly
terse handovers caused confusion and reduced diversity, often regressing
toward the mean rather than expanding the hypothesis space.

\hypertarget{opus-codegen-and-opus-image-reasoning-discarded-from-final-mix}{%
\subsection{Opus codegen and Opus image reasoning (discarded from final
mix)}\label{opus-codegen-and-opus-image-reasoning-discarded-from-final-mix}}

In the final system, Opus contributes only a single text-reasoning
candidate. Opus codegen and image reasoning were tested but contributed
less uniquely relative to the GPT/Gemini configurations, and were
dropped from the final candidate mix.

\hypertarget{grid-representations-and-output-constraints-discarded-variants}{%
\subsection{Grid representations and output constraints (discarded
variants)}\label{grid-representations-and-output-constraints-discarded-variants}}

Key findings:

\begin{itemize}
\tightlist
\item
  \textbf{CSV-style} encoding outperformed many alternatives, especially
  as grids grow and representation consumes a large share of the
  reasoning budget.
\item
  Forcing strict outputs (e.g., \textbf{requiring JSON} via API-level
  response formats) underperformed. This is important because strict
  schemas are a common LLM engineering practice, but appear to reduce
  model effectiveness on this domain.
\end{itemize}

Engineering trade-off: - removing constraints increases output noise;
robust parsing (regex + validation) becomes necessary, but was worth it
for accuracy.

\hypertarget{synthetic-data-augmentation-for-code-candidates-discarded}{%
\subsection{Synthetic data augmentation for code candidates
(discarded)}\label{synthetic-data-augmentation-for-code-candidates-discarded}}

Motivation: - generate additional training examples (e.g., via color
permutation, rotation, or mirroring of the provided pairs) to give
code-generation candidates more test cases to validate against,
potentially improving program correctness.

Reasons for discarding: - \textbf{Surface-level augmentations add little
signal.} Color permutations produce nominally distinct examples but do
not test new structural properties of the transformation; a program that
overfits to the original examples will typically also pass
color-permuted variants. - \textbf{Geometric transforms break
semantics.} Rotation and mirroring alter the meaning of tasks that
depend on absolute orientation --- for example, gravity-based tasks
(where ``down'' matters) or tasks where spatial relationships across
training pairs encode the rule (e.g., map reconstruction). Applying
these transforms indiscriminately would introduce incorrect training
signal. - \textbf{Meaningful augmentation requires solving the task
first.} Generating genuinely informative synthetic examples (new inputs
paired with correct outputs) requires knowing the transformation rule
--- which is the problem the solver is trying to solve. This makes
meaningful augmentation infeasible in a private-dataset evaluation
setting where ground-truth transformations are unavailable.

\hypertarget{extensive-prompt-engineering-discarded}{%
\subsection{Extensive prompt engineering
(discarded)}\label{extensive-prompt-engineering-discarded}}

This is perhaps the most counterintuitive finding in the paper, and it
directly contradicts standard LLM engineering practice.

The conventional approach to LLM integration treats the model as a
programmable API: the more precisely you specify the desired behavior
--- step-by-step instructions, prescribed reasoning templates, output
schemas, chain-of-thought scaffolding --- the better the results. This
works well for structured tasks like data extraction, classification, or
format conversion, where the solution space is well-defined and the
model's job is to comply with a specification.

On ARC-AGI, this approach consistently degraded performance. During
development, I tested numerous prompt engineering strategies including:

\begin{itemize}
\tightlist
\item
  \textbf{Prescribed reasoning templates:} instructing the model to
  first identify objects, then describe transformations, then apply them
  step by step.
\item
  \textbf{Structured output requirements:} requiring specific output
  formats (e.g., JSON grids via API-level response format constraints).
\item
  \textbf{Detailed chain-of-thought scaffolding:} breaking the reasoning
  into named stages (``Step 1: Identify the pattern. Step 2: Describe
  the rule. Step 3: Apply to test input.'').
\item
  \textbf{Domain-specific heuristics in the prompt:} suggesting the
  model look for symmetry, rotation, color mapping, or other common ARC
  patterns.
\item
  \textbf{Iterative prompt refinement:} tuning prompt wording based on
  failure analysis of specific tasks.
\end{itemize}

In every case, the more prescriptive the prompt, the worse the system
performed on the hardest tasks. The final system uses a deliberately
minimal prompt (Section 4) that provides only the task data, a brief
context sentence, and a request to explain reasoning --- with no
prescribed structure, no step-by-step template, and no domain
heuristics.

The mechanism appears to be a \textbf{compliance tax on reasoning}: when
the model is given detailed instructions about \emph{how} to think, it
allocates a significant portion of its reasoning budget to following
those instructions rather than to actually solving the problem. On easy
tasks --- where the transformation is simple enough that any reasonable
approach works --- this overhead is harmless. On hard tasks --- where
the model needs to make creative leaps, entertain unusual hypotheses, or
reason about structures it has never seen --- the overhead is fatal. The
model dutifully follows the prescribed template, produces a
well-formatted but wrong answer, and never explores the unstructured
reasoning path that might have led to the correct solution.

This also interacts with diversity: a prescriptive prompt narrows the
hypothesis space across candidates. When all N candidates follow the
same reasoning template, they tend to converge on the same (possibly
wrong) answer. A minimal prompt allows different candidates to approach
the problem from genuinely different angles, increasing the probability
that at least one candidate finds the correct transformation.

The implication is that for novel reasoning tasks --- where the solution
is not known in advance and the model must discover it --- \textbf{the
best prompt is often the least prescriptive one}. The engineer's job
shifts from programming the model's behavior to removing obstacles to
the model's reasoning. This is uncomfortable for systems engineers
accustomed to treating prompts as specifications, but on this benchmark,
letting the model think freely and tolerating noisy outputs (with robust
downstream parsing) consistently outperformed carefully engineered
prompts.

\begin{center}\rule{0.5\linewidth}{0.5pt}\end{center}

\hypertarget{discussion-and-conclusion}{%
\section{Discussion and Conclusion}\label{discussion-and-conclusion}}

This paper demonstrates that strong ARC-AGI-2 performance can be
achieved by treating \textbf{modalities as search operators} and pairing
diverse candidate generation with context-preserving selection:

\begin{itemize}
\tightlist
\item
  generate candidates independently across heterogeneous reasoning
  channels (text, image, code) to maximize hypothesis diversity,
\item
  then select using holistic judging over full traces.
\end{itemize}

At the time of writing, this approach achieves the highest score on the
ARC Prize Verified leaderboard (72.9\%), surpassing the best-performing
standalone frontier models --- GPT-5.2 Pro (54.2\%) and Gemini 3 Pro
(54.0\%) --- by +18.7 percentage points. This substantial margin
suggests that orchestrating diverse reasoning modalities with principled
selection is a more effective strategy for abstract reasoning than
scaling any single model alone.

\hypertarget{limitations}{%
\subsection{Limitations}\label{limitations}}

This work has several important limitations that should be weighed when
interpreting the results.

\textbf{Cost and practical scalability.} The system spends \$19.69 per
task on the public evaluation and \$38.99 per task on the semi-private
set --- orders of magnitude more expensive than a single model call. ARC
Prize explicitly treats efficiency as a first-class metric (Chollet et
al. 2024), and the current system is far from efficient. The brute-force
strategy of generating 29 candidates across three models and three
modalities, then running three separate judge passes over the full
candidate pool, is effective but wasteful. Much of the cost is spent on
candidates that contribute nothing to the final answer. A production
system would need adaptive routing --- spending heavily only on tasks
that resist cheap methods --- but no such routing mechanism has been
developed here.

\textbf{Single-run results.} The headline numbers (76.11\% public,
72.9\% semi-private) each come from a single evaluation run. LLM outputs
are stochastic, and the system's reliance on sampling diversity means
that results will vary across runs. No confidence intervals are reported
because repeated full-pipeline runs were not performed (each costing
\textasciitilde\$2,400). The true expected accuracy could be
meaningfully higher or lower than the reported figures, and the variance
is unknown.

\textbf{Incomplete ablation coverage.} Section 7 reports post-hoc
ablations for holistic selection (+7 instances) and synthesis (+1
instance), but these are extracted from a single run rather than from
controlled experiments with matched baselines. Several ablations that
would substantially strengthen the paper's claims --- end-to-end
modality removal, independent vs sequential generation, candidate budget
scaling, judge ensemble sizing --- have not been performed due to cost
constraints. The component attribution claims in this paper should be
understood as indicative rather than rigorous. Section 7 provides a
detailed list of unperformed ablations.

\textbf{Reproducibility fragility.} The system depends on specific
proprietary model snapshots (GPT-5.2, Gemini 3 Preview, Opus 4.5) and
vendor-specific ``reasoning settings'' (e.g., OpenAI's ``x-high''
reasoning effort) that may change or become unavailable over time. Model
providers routinely update model weights, deprecate API parameters, and
alter rate limits without notice. Results may not be exactly
reproducible even with identical prompts and parameters if the
underlying model versions drift. To maximize reproducibility within
these constraints, the full source code is open-sourced at
https://github.com/beetree/ARC-AGI and the complete raw data for the
public evaluation run --- including all prompts, responses, reasoning
traces, and judge transcripts (over 7 million lines) --- is available on
Kaggle at
https://www.kaggle.com/code/johanland/johan-land-solver-v7-public/comments?scriptVersionId=290052212.
Exact replication is nonetheless not guaranteed.

\textbf{No learning across tasks.} The system treats each task
independently --- no information is carried from one task to the next. A
human solver would build intuitions across tasks (e.g., ``tasks in this
benchmark often involve symmetry'' or ``I've seen this color-mapping
pattern before''), but the current system starts from scratch every
time. This is both a limitation and a design choice: task independence
simplifies the pipeline and avoids overfitting to task ordering, but it
means the system cannot amortize its reasoning cost across related
tasks.

\textbf{Narrow evaluation domain.} The results are demonstrated on a
single benchmark (ARC-AGI-2). While the architectural pattern ---
diverse generation plus holistic judging --- is domain-general in
principle, this paper provides no evidence that the approach transfers
to other domains. The specific design choices (modality mix, candidate
count, judge prompt structure) were tuned for ARC and may not generalize
without adaptation.

\hypertarget{future-work}{%
\subsection{Future work}\label{future-work}}

\begin{itemize}
\tightlist
\item
  Adaptive routing: allocate expensive modalities only when uncertainty
  is high.
\item
  Judge compression without premature abstraction: find ways to reduce
  context size while retaining the benefits of joint context.
\item
  Further ablations: Section 7 lists a detailed set of unperformed
  ablations --- including judge ensemble sizing, trace content
  contribution, per-model attribution, and early-stopping threshold
  tuning --- that would strengthen component attribution claims.
\item
  Synthesis gating and amplification: the current judge always has the
  option to synthesize a novel output. A gating mechanism that decides
  \emph{when} synthesis is likely to help --- e.g., only when candidate
  agreement is low or when no candidate passes a confidence threshold
  --- could improve targeting. Conversely, when synthesis is identified
  as having high potential (e.g., multiple candidates contain
  complementary partial solutions), the system could invoke additional
  synthesis attempts with varied prompting to increase the probability
  of a correct recombination. The current single-pass synthesis yielded
  +1 instance (Section 7); a more aggressive, targeted synthesis
  strategy could yield further uplift on the hardest tasks where no
  single candidate is fully correct.
\item
  Image representation tuning: the finding that intentionally imprecise
  grid renderings outperform pixel-perfect ones (Section 4) is
  suggestive but not well understood. Systematic study of rendering
  parameters --- resolution, distortion level, color palette, annotation
  style --- and their interaction with different vision-language models
  could yield further gains and clarify when and why visual prompting
  helps.
\item
  Broader modality coverage: additional frontier providers and
  open-source models, plus parameter sweeps (temperature, etc.).
\item
  Formal diversity quantification: the current paper measures modality
  complementarity via oracle overlap counts (Tables 3--4), but a richer
  diversity measure --- e.g., pairwise output disagreement rates,
  embedding-space distances between reasoning traces, or
  information-theoretic metrics over the candidate distribution ---
  would enable principled decisions about which generators to add,
  remove, or scale. A task-archetype taxonomy (classifying ARC tasks by
  the type of reasoning they require) could further clarify which
  modalities are most valuable for which problem classes.
\item
  Domain transfer: the ``diverse generation + holistic judging'' pattern
  is not specific to ARC. Any domain where models produce confident but
  divergent answers --- mathematical proof search, legal analysis,
  medical diagnosis --- could benefit from context-preserving
  adjudication over multiple independent reasoning traces. Validating
  this on non-ARC benchmarks is a natural next step.
\end{itemize}

\hypertarget{a-note-on-ai-assisted-development}{%
\subsection{A note on AI-assisted
development}\label{a-note-on-ai-assisted-development}}

The solver was developed with AI assistance for both implementation and
design (Section 3.3), with the author directing strategy and evaluating
results. That this approach produced a competitive result in a
solo-researcher setting suggests that AI-assisted development is a
practical paradigm for complex pipelines.

\hypertarget{reproducibility}{%
\subsection{Reproducibility}\label{reproducibility}}

The full source code for the solver is available at:
https://github.com/beetree/ARC-AGI. The repository contains all prompts,
tool schemas, candidate generation configurations, and judging logic.

The complete public-evaluation run --- including all API parameters,
model versions, and raw logs (prompts, responses, reasoning traces,
intermediate artifacts, and judge transcripts; over 7 million lines) ---
is available as a Kaggle notebook:
https://www.kaggle.com/code/johanland/johan-land-solver-v7-public/comments?scriptVersionId=290052212.

The semi-private evaluation was executed by ARC Prize's verification
infrastructure; the author does not control that environment and cannot
release those logs.

\hypertarget{acknowledgments}{%
\section{Acknowledgments}\label{acknowledgments}}

Thank you to the ARC-AGI Discord community for valuable discussion and
shared insights throughout the development of this work, and to Greg
Kamradt at ARC Prize for conducting the official semi-private
evaluation.

\hypertarget{refs}{}
\begin{CSLReferences}{1}{0}
\leavevmode\vadjust pre{\hypertarget{ref-akyurek2024surprising}{}}%
Akyürek, Ekin, Mehul Damani, Adam Zweiger, Linlu Qiu, Han Guo, Jyothish
Pari, Yoon Kim, and Jacob Andreas. 2025. {``The Surprising Effectiveness
of Test-Time Training for Few-Shot Learning.''} In \emph{Proceedings of
the 42nd International Conference on Machine Learning}, 267:942--63.
PMLR.

\leavevmode\vadjust pre{\hypertarget{ref-besta2024graph}{}}%
Besta, Maciej, Nils Blach, Ales Kubicek, Robert Gerstenberger, Michal
Podstawski, Lukas Gianinazzi, Joanna Gajber, et al. 2024. {``Graph of
Thoughts: Solving Elaborate Problems with Large Language Models.''} In
\emph{Proceedings of the AAAI Conference on Artificial Intelligence}.
Vol. 38.

\leavevmode\vadjust pre{\hypertarget{ref-chollet2019measure}{}}%
Chollet, François. 2019. {``On the Measure of Intelligence.''}
\emph{arXiv Preprint arXiv:1911.01547}.

\leavevmode\vadjust pre{\hypertarget{ref-arcprize2024report}{}}%
Chollet, François, Mike Knoop, Gregory Kamradt, and Bryan Landers. 2024.
{``{ARC Prize 2024}: Technical Report.''} \emph{arXiv Preprint
arXiv:2412.04604}.

\leavevmode\vadjust pre{\hypertarget{ref-ferre2021arc}{}}%
Ferré, Sébastien. 2021. {``First Steps of an Approach to the {ARC}
Challenge Based on Descriptive Grid Models and the Minimum Description
Length Principle.''} \emph{arXiv Preprint arXiv:2112.00848}.

\leavevmode\vadjust pre{\hypertarget{ref-fletcherhill2024miniarc}{}}%
Fletcher-Hill, Paul. 2024. {``Mini-{ARC}: Solving Abstraction and
Reasoning Puzzles with Small Transformer Models.''}

\leavevmode\vadjust pre{\hypertarget{ref-gao2023pal}{}}%
Gao, Luyu, Aman Madaan, Shuyan Zhou, Uri Alon, Pengfei Liu, Yiming Yang,
Jamie Callan, and Graham Neubig. 2023. {``{PAL}: Program-Aided Language
Models.''} \emph{International Conference on Machine Learning}.

\leavevmode\vadjust pre{\hypertarget{ref-hodel2024rearc}{}}%
Hodel, Michael. 2024. {``Addressing the Abstraction and Reasoning Corpus
via Procedural Example Generation.''} \emph{arXiv Preprint
arXiv:2404.07353}.

\leavevmode\vadjust pre{\hypertarget{ref-li2024induction}{}}%
Li, Wen-Ding, Keya Hu, Carter Larsen, Yuqing Wu, Simon Alford, Caleb
Woo, Spencer M. Dunn, et al. 2025. {``Combining Induction and
Transduction for Abstract Reasoning.''} In \emph{International
Conference on Learning Representations}.

\leavevmode\vadjust pre{\hypertarget{ref-bonnet2024latent}{}}%
Macfarlane, Matthew V., and Clément Bonnet. 2025. {``Searching Latent
Program Spaces.''} In \emph{Advances in Neural Information Processing
Systems}. Vol. 38.

\leavevmode\vadjust pre{\hypertarget{ref-madaan2023selfrefine}{}}%
Madaan, Aman, Niket Tandon, Prakhar Gupta, Skyler Hallinan, Luyu Gao,
Sarah Wiegreffe, Uri Alon, et al. 2023. {``Self-Refine: Iterative
Refinement with Self-Feedback.''} \emph{Advances in Neural Information
Processing Systems} 36.

\leavevmode\vadjust pre{\hypertarget{ref-moffitt2025arcgen}{}}%
Moffitt, Michael D. 2025. {``{ARC-GEN}: A Mimetic Procedural Benchmark
Generator for the Abstraction and Reasoning Corpus.''} \emph{arXiv
Preprint arXiv:2511.00162}.

\leavevmode\vadjust pre{\hypertarget{ref-moskvichev2023conceptarc}{}}%
Moskvichev, Arseny, Victor Vikram Odouard, and Melanie Mitchell. 2023.
{``The {ConceptARC} Benchmark: Evaluating Understanding and
Generalization in the {ARC} Domain.''} \emph{Transactions on Machine
Learning Research}.

\leavevmode\vadjust pre{\hypertarget{ref-ouellette2024neurally}{}}%
Ouellette, Simon. 2024. {``Towards Efficient Neurally-Guided Program
Induction for {ARC-AGI}.''} \emph{arXiv Preprint arXiv:2411.17708}.

\leavevmode\vadjust pre{\hypertarget{ref-puget2024nGPT}{}}%
Puget, Jean-François. 2024. {``A {2D} {nGPT} Model for {ARC} Prize.''}

\leavevmode\vadjust pre{\hypertarget{ref-schick2023toolformer}{}}%
Schick, Timo, Jane Dwivedi-Yu, Roberto Dessì, Roberta Raileanu, Maria
Lomeli, Luke Zettlemoyer, Nicola Cancedda, and Thomas Scialom. 2023.
{``Toolformer: Language Models Can Teach Themselves to Use Tools.''} In
\emph{Advances in Neural Information Processing Systems}. Vol. 36.

\leavevmode\vadjust pre{\hypertarget{ref-shinn2023reflexion}{}}%
Shinn, Noah, Federico Cassano, Ashwin Gopinath, Karthik Narasimhan, and
Shunyu Yao. 2023. {``Reflexion: Language Agents with Verbal
Reinforcement Learning.''} \emph{Advances in Neural Information
Processing Systems} 36.

\leavevmode\vadjust pre{\hypertarget{ref-snell2024scaling}{}}%
Snell, Charlie, Jaehoon Lee, Kelvin Xu, and Aviral Kumar. 2025.
{``Scaling {LLM} Test-Time Compute Optimally Can Be More Effective Than
Scaling Model Parameters.''} In \emph{International Conference on
Learning Representations}.

\leavevmode\vadjust pre{\hypertarget{ref-wang2022selfconsistency}{}}%
Wang, Xuezhi, Jason Wei, Dale Schuurmans, Quoc Le, Ed Chi, Sharan
Narang, Aakanksha Chowdhery, and Denny Zhou. 2023. {``Self-Consistency
Improves Chain of Thought Reasoning in Language Models.''} In
\emph{International Conference on Learning Representations}.

\leavevmode\vadjust pre{\hypertarget{ref-wei2022chain}{}}%
Wei, Jason, Xuezhi Wang, Dale Schuurmans, Maarten Bosma, Brian Ichter,
Fei Xia, Ed Chi, Quoc Le, and Denny Zhou. 2022. {``Chain-of-Thought
Prompting Elicits Reasoning in Large Language Models.''} \emph{Advances
in Neural Information Processing Systems} 35.

\leavevmode\vadjust pre{\hypertarget{ref-yao2023tree}{}}%
Yao, Shunyu, Dian Yu, Jeffrey Zhao, Izhak Shafran, Thomas L. Griffiths,
Yuan Cao, and Karthik Narasimhan. 2023. {``Tree of Thoughts: Deliberate
Problem Solving with Large Language Models.''} In \emph{Advances in
Neural Information Processing Systems}. Vol. 36.

\leavevmode\vadjust pre{\hypertarget{ref-yao2022react}{}}%
Yao, Shunyu, Jeffrey Zhao, Dian Yu, Nan Du, Izhak Shafran, Karthik
Narasimhan, and Yuan Cao. 2023. {``{ReAct}: Synergizing Reasoning and
Acting in Language Models.''} \emph{International Conference on Learning
Representations}.

\leavevmode\vadjust pre{\hypertarget{ref-zheng2023judging}{}}%
Zheng, Lianmin, Wei-Lin Chiang, Ying Sheng, Siyuan Zhuang, Zhanghao Wu,
Yonghao Zhuang, Zi Lin, et al. 2023. {``Judging {LLM}-as-a-Judge with
{MT-Bench} and {Chatbot Arena}.''} \emph{Advances in Neural Information
Processing Systems} 36.

\end{CSLReferences}

\end{document}